\documentclass[pdflatex,sn-mathphys-num,iicol]{sn-jnl}

\usepackage{graphicx}%
\usepackage{multirow}%
\usepackage{amsmath,amssymb,amsfonts}%
\usepackage{amsthm}%
\usepackage{mathrsfs}%
\usepackage[title]{appendix}%
\usepackage{xcolor}%
\usepackage{textcomp}%
\usepackage{manyfoot}%
\usepackage{booktabs}%
\usepackage{algorithm}%
\usepackage{algorithmicx}%
\usepackage{algpseudocode}%
\usepackage{listings}%
\usepackage{makecell}

\theoremstyle{thmstyleone}%
\theoremstyle{thmstyletwo}%

\theoremstyle{thmstylethree}%

\begin{document}

\title[Event-Based Early Warning of Vineyard Disease Risk from Environmental Time Series]{Event-Based Early Warning of Vineyard Disease Risk from Environmental Time Series}


\author*[1]{\fnm{Ivica} \sur{Dimitrovski}}\email{ivica.dimitrovski@finki.ukim.mk}

\author[1]{\fnm{Ivan} \sur{Kitanovski}}\email{ivan.kitanovski@finki.ukim.mk}

\author[1]{\fnm{Danco} \sur{Davcev}}\email{danco.davcev@finki.ukim.mk}

\author[1]{\fnm{Slobodan} \sur{Kalajdziski}}\email{slobodan.kalajdziski@finki.ukim.mk}

\author[1]{\fnm{Kosta} \sur{Mitreski}}\email{kosta.mitreski@finki.ukim.mk}

\affil[1]{\orgdiv{Faculty of Computer Science and Engineering}, \orgname{Ss. Cyril and Methodius University}, \orgaddress{\street{ul.Rudzer Boshkovikj 16, P.O. 393}, \city{Skopje}, \postcode{1000}, \country{North Macedonia}}}


\abstract{Accurate early warning of vineyard disease risk can support timely intervention and more targeted crop-protection decisions. However, conventional daily-status classification does not directly measure whether a model can anticipate the beginning of a new risk episode. This study investigates an event-based formulation for short-horizon warning of vineyard mildew risk from agro-meteorological time series. The available downy- and powdery-mildew annotations are derived from fungicide-treatment records and are therefore interpreted as management-oriented risk proxies rather than direct pathological observations. Because the two labels agree on 98.9\% of observed days and are identical in the validation and test years, they are combined into an operational any-mildew target without implying epidemiological equivalence. New events are defined after a minimum disease-free gap, and models estimate whether an event will begin within the following 3--7 days. Using a chronological 2020--2021/2022/2023 train-validation-test split, we compare calendar-only, environmental-only, and combined representations with logistic regression, XGBoost, Long Short-Term Memory (LSTM) networks, and Temporal Convolutional Networks (TCNs). Calendar-only models were highly competitive, with calendar-based LSTM and TCN models detecting 7 of 8 test events, while a monthly climatological baseline detected 6 events. Environmental-only models were substantially weaker, and adding environmental variables to calendar information produced inconsistent gains across model families. A comparison with conventional daily-status classification showed that the two formulations capture different objectives and exhibit different trade-offs in point-wise performance, event coverage, and alert burden, without uniform superiority of either formulation. Sensitivity analyses further showed that results depend on the historical window, prediction horizon, and event-separation gap. SHapley Additive exPlanations (SHAP) analysis indicated that the fitted model used both seasonal and meteorological information, without revealing a single dominant or consistently interpretable environmental precursor. These findings demonstrate the value of event-oriented evaluation for assessing anticipatory warning and emphasize the need for explicit seasonal baselines and cautious biological interpretation of management-derived proxy targets.}

\keywords{event-based early warning, grapevine disease warning, daily-status classification, treatment-derived proxy labels, seasonal confounding, agro-meteorological time series}



\maketitle

\section{Introduction}
Grapevine cultivation is one of the most economically important agricultural activities worldwide, yet vineyard productivity and grape quality remain highly vulnerable to major diseases such as downy mildew and powdery mildew \citep{velasquez2022current}. The development of these diseases is influenced by environmental conditions including temperature, humidity, rainfall, and leaf wetness, making viticulture a natural setting for predictive modelling from agro-meteorological observations. At the same time, increasing pressure to reduce pesticide use and improve the efficiency of crop protection has intensified the need for warning systems that support timely and targeted interventions \citep{steffenel2023ai}.

Traditionally, vineyard disease management has relied on calendar-based spraying schedules, expert-defined rules, and weather-driven infection models. More recently, the field has expanded toward digital and data-driven viticulture through IoT-enabled meteorological monitoring, machine learning on agro-meteorological variables, remote and proximal sensing, and multimodal disease-risk assessment \citep{portela2024systematic,velasquez2022current,kotaidou2026grapevine}. This development is consistent with broader AI--IoT precision-agriculture frameworks that integrate sensing, intelligent decision support, edge or cloud processing, and autonomous or semi-autonomous agricultural workflows \citep{davcev2026agentic}. These approaches aim to move crop protection from static preventive schedules toward adaptive decisions informed by continuously collected observations.

Despite this progress, many existing studies formulate vineyard disease prediction as daily disease-state classification \citep{arvanitis2025utilizing}, risk-level estimation \citep{kotaidou2026grapevine}, severe-attack forecasting \citep{chen2020forecasting}, first-appearance prediction for a specific pathogen or cultivar \citep{valori2023advanced}, or image- and sensor-based disease classification \citep{rocscuaneanu2022detection}. These formulations address different operational objectives and should not be treated as interchangeable. In particular, strong daily-status performance does not necessarily demonstrate successful anticipation of a new warning-relevant episode. When positive annotations persist over several consecutive days, a model may obtain favorable point-wise scores by reproducing an already active state rather than by predicting the transition into that state. Consequently, conventional classification metrics alone may overstate the practical early-warning capability of a model.

A second methodological issue concerns the provenance and meaning of the target annotations. In operational agricultural datasets, disease-related labels may be derived from fungicide-treatment records, expert management decisions, or preventive actions rather than from direct pathological confirmation. Such proxy annotations can reflect recent environmental conditions, but may also encode spraying schedules, phenological timing, grower practices, and other recurring seasonal patterns. This raises a central question for data-driven warning systems: to what extent do predictive models use short-term environmental information, and to what extent do they exploit seasonal or management-related regularities? The question is particularly important when month and day-of-year variables are included explicitly as model inputs.

Motivated by these limitations, this study formulates vineyard disease-risk forecasting as an event-based early-warning problem \citep{terada2025realizing}. Instead of predicting the daily risk status, the task is to determine whether a new annotated risk episode will begin within a future interval of 3--7 days. A new event is defined as the first positive day following a minimum disease-free gap, which reduces fragmentation caused by short interruptions in the binary sequence. The event formulation therefore focuses on transitions into new risk periods rather than persistence of the current status. It is not assumed, however, that the resulting events correspond to pathologically confirmed infection onset or that successful prediction necessarily demonstrates the identification of biological disease precursors.

The study uses multi-year agro-meteorological observations together with downy- and powdery-mildew annotations derived from fungicide-treatment records \citep{arvanitis2025utilizing}. Although the two diseases have distinct and partly contrasting epidemiological drivers, the available annotations agree on 98.9\% of observed days and are identical in the validation and test years. They are therefore combined into an operational any-mildew target. This aggregation is motivated by the structure of the available annotations and the limited number of independent events, and should not be interpreted as implying epidemiological equivalence between downy and powdery mildew. The resulting target is treated as a management-oriented disease-risk proxy relevant to vineyard decision support rather than as a pathogen-specific pathological outcome.

To examine which information supports prediction, three complementary input representations are compared: calendar-only features, environmental features without explicit calendar information, and their combination. This comparison is necessary because treatment-associated events may occur at recurring times of the year, allowing apparently strong prediction even in the absence of short-term meteorological information. A monthly climatological baseline is additionally used to quantify how much event-warning performance can be obtained from recurring seasonal timing alone. The event-onset formulation is also compared with conventional daily-status classification using the same chronological data split and representative model families. This comparison examines whether favorable point-wise classification performance corresponds to useful warning of new episodes when predictions are assessed under a common event-oriented protocol.

The predictive models include logistic regression as an interpretable baseline, XGBoost as a strong tabular learner \citep{chen2016xgboost}, Long Short-Term Memory networks as recurrent sequence models \citep{siami2019performance}, and Temporal Convolutional Networks as convolutional temporal models \citep{hewage2020temporal}. These models provide complementary linear, tree-based, recurrent, and convolutional perspectives under the same chronological protocol. The objective is not to introduce a new predictive architecture, but to investigate how conclusions change with target formulation, input representation, temporal design, and model family.

All models are evaluated using a strictly chronological train-validation-test split, with 2020--2021 used for training, 2022 for validation and model selection, and 2023 reserved for final testing. Sample-level metrics including F1-score, Area Under the Receiver Operating Characteristic Curve (AUROC), and area under the precision--recall curve (PR-AUC) are complemented by an event-oriented evaluation protocol. The latter measures whether true events are anticipated within the valid warning interval, how much lead time is provided, how daily alerts are grouped into episodes, and how slightly mistimed alerts are distinguished from strict false alerts. Sensitivity analyses are conducted for the historical window length, prediction horizon, and disease-free event gap. Model interpretation is further supported through SHapley Additive exPlanations (SHAP) analysis~\cite{lundberg2017unified}, which is used to examine how the representative fitted model distributes attribution across seasonal and environmental feature groups and historical lags.

Accordingly, the study addresses the following questions:

\begin{itemize}
    \item To what extent are treatment-associated mildew-risk events predictable from cyclic calendar information, environmental observations, or their combination?

    \item How does event-onset prediction differ from conventional daily-status classification in terms of point-wise performance, onset-event coverage, warning lead time, and alert burden?

    \item How robust are the conclusions to disease-target construction, historical window length, prediction horizon, disease-free event gap, model family, and random initialization?

    \item How does a representative fitted XGBoost model distribute attribution across seasonal and environmental feature groups and historical lags?
\end{itemize}

The methodological contribution of this work lies primarily in the formulation and evaluation of the prediction problem rather than in the predictive models themselves. The main contributions are as follows:

\begin{itemize}
    \item We formulate vineyard disease-risk prediction as short-horizon warning of new treatment-associated risk episodes rather than prediction of a persistent daily status.

    \item We analyse the provenance, temporal structure, and agreement of the original downy- and powdery-mildew annotations, clarifying the operational meaning and biological limitations of the unified any-mildew target.

    \item We compare a monthly climatological baseline together with calendar-only, environmental-only, and combined representations, showing that seasonal timing accounts for a substantial proportion of predictive performance, while the change in predictive performance after adding environmental variables varies across model families.

    \item We contrast event-onset prediction with conventional daily-status classification and evaluate both formulations under a common event-oriented protocol, showing that point-wise classification performance, onset-event coverage, and alert burden represent related but distinct aspects of model behavior.

    \item We evaluate logistic regression, XGBoost, LSTM, and TCN models under a chronological, multi-seed protocol and show that model rankings depend on the input representation and temporal task definition.

    \item We assess sensitivity to the historical window length, prediction horizon, and event-separation gap, demonstrating that these parameters materially influence sample construction, event counts, and model conclusions.

    \item We introduce a precisely defined event-oriented evaluation protocol covering event coverage, warning lead time, alert burden, episode formation, and near-miss-aware identification of strict false alerts.

    \item We use SHAP analysis to examine how models use seasonal and environmental information while distinguishing predictive association from biological causal interpretation.
\end{itemize}

Overall, this work shows that event-based evaluation provides a useful perspective for distinguishing warning of new transitions from classification of persistent daily risk status. At the same time, the results demonstrate that treatment-derived vineyard annotations contain substantial seasonal and management-related structure and therefore require explicit calendar baselines and cautious interpretation. The study is presented as a single-site temporal case study, and its findings should not be interpreted as evidence of broad cross-vineyard generalization, direct prediction of pathological disease onset, or statistical superiority of a particular model family.

The remainder of this paper is organized as follows. Section~2 reviews related work on vineyard disease prediction, environmental time-series modelling, proxy-labelled risk warning, and event-oriented evaluation. Section~3 presents the dataset, target construction, preprocessing, feature representations, and predictive models. Section~4 describes the chronological experimental protocol, baseline formulations, sensitivity analyses, and evaluation measures. Section~5 reports the predictive, event-oriented, sensitivity, and interpretation results. Finally, Section~6 summarizes the principal findings, limitations, and directions for future work.

\section{Related Work}

\subsection{Vineyard Disease Warning from Environmental Data}

Vineyard disease warning has traditionally relied on empirical thresholds, expert-defined rules, and mechanistic infection-risk models based on plant-pathology knowledge. These systems typically use environmental variables such as temperature, relative humidity, rainfall, and leaf wetness to estimate whether conditions are favorable for infection or disease development \citep{velasquez2022current}. Their main advantages are interpretability and direct agronomic relevance. However, they often require local calibration and may not transfer directly across vineyards, cultivars, seasons, and sensing configurations \citep{trilles2019adapting,mezei2022grapevine}.

The increasing availability of IoT weather stations and continuous agro-meteorological monitoring has encouraged data-driven approaches~\cite{sannakki2013neural},~\cite{chavan2019iot}. Machine-learning methods have been applied to severe downy-mildew attack forecasting \citep{chen2020forecasting}, downy-mildew infection-event prediction \citep{steffenel2023ai}, powdery-mildew first-appearance forecasting \citep{valori2023advanced}, and risk assessment from atmospheric and sensor observations \citep{marcu2022predictive,trilles2019adapting}. Other studies combine environmental measurements with imagery, proximal sensing, or multimodal data to estimate disease-related risk \citep{portela2024systematic,kotaidou2026grapevine}. More broadly, semantic segmentation supports the extraction of spatially explicit information from aerial and satellite imagery for agricultural and environmental monitoring \cite{spasev2023semantic}, while deep multimodal fusion enables complementary remote-sensing sources to be integrated within unified analysis frameworks \cite{dimitrovski2024deep}. These spatial and multimodal approaches address a different problem setting from the present station-based time-series study, but they indicate how future vineyard warning systems could combine temporal environmental risk estimates with spatial observations. Collectively, the reviewed works demonstrate the value of environmental and multimodal monitoring for vineyard decision support, while also addressing different objectives, including current-state classification, risk-level estimation, severity forecasting, first appearance, and infection-event warning.

Of particular relevance, Arvanitis et al. \citep{arvanitis2025utilizing} introduced the dataset used in the present study and formulated the task as daily classification of grapevine disease-related labels from IoT environmental observations. Their work demonstrated the feasibility of applying advanced tabular learning to this dataset. The present study revisits the same data from a different operational perspective, asking whether models can anticipate the beginning of a new treatment-associated risk episode rather than classify the daily status. This distinction is important because positive annotations persist over multiple consecutive days, and favorable point-wise performance may therefore reflect continuity of an already active state rather than successful onset warning.

Another important distinction concerns label provenance. Vineyard datasets may use direct pathological observations, visual symptom assessments, expert risk scores, mechanistic infection indicators, or treatment records as prediction targets. These forms of supervision are not equivalent. Fungicide-treatment records can provide useful operational proxies, but they may reflect preventive schedules, phenological timing, grower experience, logistical constraints, and prior field observations in addition to environmental disease risk. Models trained on such annotations may therefore exploit management-related or seasonal regularities. This makes explicit calendar baselines particularly important when evaluating treatment-derived labels.

The dataset examined in this work belongs to this proxy-labelled category. Its downy- and powdery-mildew annotations are derived from treatment-associated records rather than direct pathological confirmation. Moreover, the two labels show almost complete temporal agreement. The present study therefore interprets their union as an operational any-mildew risk proxy, without implying that the two pathogens have equivalent epidemiological requirements or that the resulting target supports pathogen-specific biological conclusions.

\subsection{Machine Learning for Agricultural and Environmental Risk Warning}

Agricultural and environmental early-warning problems are commonly characterized by nonlinear relationships, correlated predictors, class imbalance, temporal dependence, and a limited number of independent events. In such settings, classical methods such as logistic regression, random forests, and gradient-boosted trees remain important because they provide strong baselines, can operate effectively on structured descriptors, and generally require fewer observations than high-capacity deep architectures. Comparative crop-disease studies have shown that tree ensembles and regularized models can remain competitive with neural networks when meteorological data are limited and highly structured \citep{fenu2021forecasting,bijlwan2025predicting}. In vineyard applications, classical machine-learning models have also produced useful results for downy-mildew warning from agro-meteorological station data \citep{steffenel2023ai}.

Feature construction is particularly important in environmental-risk modelling. Biological and environmental processes are often influenced not by isolated daily measurements, but by accumulation, persistence, interactions, and seasonal context. Accordingly, predictive representations may include rainfall sums, sustained-humidity indicators, temperature ranges, counts of favorable days, temporal interactions, or seasonally crossed features.

Broader agricultural and environmental studies reinforce this point. Mechanism-guided risk warning has been explored for grasshopper outbreaks~\cite{su2026grasshopperml}, while automatically constructed multispectral features have been used for banana Fusarium-wilt recognition~\cite{su2025banana}. In forest-fire risk prediction, feature crosses, model fusion, multi-stage feature selection, and AutoML have been used to represent seasonal and environmental interactions more effectively \citep{su2024fc,su2024efficient}. Although these studies involve different domains and data modalities, they illustrate a common methodological principle: environmental warning depends not only on model architecture, but also on how raw observations are transformed, selected, combined, and evaluated.

Among deep learning approaches, Long Short-Term Memory (LSTM) networks are frequently used because they preserve temporal order and can represent cumulative effects across historical observations. LSTM-based components have been applied to weather-driven vineyard risk assessment and other climate-sensitive agricultural forecasting tasks \citep{kotaidou2026grapevine,veerappan2026ai,ratna2025harnessing}. Hybrid CNN--LSTM architectures have similarly been proposed to combine local pattern extraction with recurrent sequence modelling \citep{jain2021ai}.

Temporal Convolutional Networks (TCNs) provide a convolutional alternative to recurrent models. Through causal and dilated convolutions, TCNs can capture local and medium-range dependencies while supporting parallel computation and stable optimization. Although TCNs remain less common in vineyard disease warning, they have shown competitive performance in other agricultural time-series applications, including crop-yield prediction from satellite image sequences \citep{yli2022scalable}.

Recent studies have also explored transformer-related and foundation-style models for agricultural prediction. TabPFN and related tabular learners have been evaluated alongside XGBoost, CatBoost, and LightGBM for grapevine disease classification \citep{zhao2024grapevine,arvanitis2025utilizing}. However, the effective amount of independent information in small agricultural datasets is determined by the number of seasons, sites, and events rather than by the nominal number of overlapping windows. In the present setting, only four annual cycles and a small number of independent treatment-associated events are available. We therefore focus on logistic regression, XGBoost, LSTM, and TCN as representative lower-capacity linear, tree-based, recurrent, and convolutional models. The objective is not to establish a comprehensive architecture benchmark, but to examine how conclusions change with target formulation, feature representation, and evaluation protocol.

\subsection{From Daily Status Classification to Event-Based Early Warning}

Most supervised vineyard disease studies formulate the learning task as daily status classification, risk-level estimation, or severity prediction. This formulation is appropriate when the objective is contemporaneous risk assessment. However, it does not directly answer whether the beginning of a new risk period can be anticipated sufficiently early to support intervention.

The distinction becomes important when positive labels are temporally persistent. If an annotation remains active for several consecutive days, adjacent prediction samples may be highly similar, and a model can obtain strong point-wise scores by reproducing the continuation of the positive state. Such performance does not necessarily imply that the model can anticipate the transition that initiated the risk period.

Some vineyard studies move closer to transition-oriented forecasting through first-appearance prediction or infection-event modelling \citep{valori2023advanced,steffenel2023ai}. Nevertheless, direct comparison between persistent daily-status classification and prediction of new, temporally separated episodes remains uncommon. The two tasks differ in operational meaning, class balance, and difficulty. Daily-status classification identifies whether an annotation is active on an individual day, whereas event-onset prediction retains only historical windows associated with an upcoming transition into a new positive episode.

Event-oriented prediction is well established in the broader time-series literature. Early-classification research formalizes the trade-off between prediction accuracy and the amount of historical information available before a decision is made \citep{xing2009early}. Methods such as TEASER explicitly evaluate accuracy together with earliness \citep{schafer2020teaser}. Other work on temporal event prediction argues that point-wise binary classification is insufficient when the practical objective is to anticipate discrete future events \citep{yeche2023temporal}. These ideas provide a conceptual basis for distinguishing daily vineyard risk classification from short-horizon warning of a new risk episode.

In the present study, an event is defined as the first positive day following a minimum disease-free interval. The gap is used to prevent brief interruptions in the binary annotation from being treated as independent episodes. It should not be interpreted as a pathogen-specific biological incubation period. Instead, it is an operational target-construction parameter whose influence is examined through sensitivity analysis.

The event-onset formulation is compared with conventional daily-status classification using common model families and the same chronological data split. The daily formulation provides a direct representation of the point-wise task used in previous work, while the event formulation asks whether a new episode will begin within a predefined warning horizon. Because the two formulations use different targets and eligible samples, their sample-level scores are interpreted within each task rather than as directly interchangeable measures. Their predictions are additionally examined under a common event-oriented protocol to compare onset-event coverage and alert burden.

\subsection{Evaluation and Interpretation of Early-Warning Systems}

Vineyard disease studies commonly report accuracy, precision, recall, F1-score, and AUROC \citep{chen2020forecasting,arvanitis2025utilizing,kotaidou2026grapevine}. These metrics are useful for sample-level discrimination, but they do not directly indicate whether an alert is issued before a new event, how much warning time is provided, whether several daily alerts correspond to the same event, or whether event coverage is achieved through an excessive number of alerts.

Early-warning evaluation should therefore consider both sample-level and event-level behavior. Relevant quantities include event recall, warning lead time, the number of alert days, the number of distinct alert episodes, and the distinction between slightly mistimed alerts and alerts unrelated to any event. Reproducible evaluation also requires explicit rules defining how alerts are matched to events, how adjacent alerts are grouped into episodes, how multiple alerts for one event are handled, and how near-miss and strict false alerts are distinguished.

Seasonal baselines are also important when event timing follows recurring annual patterns. A monthly climatological baseline can quantify how much warning performance is obtainable from historical calendar structure without environmental observations or a fitted machine-learning model. Learned calendar-only models can then determine whether more detailed cyclic temporal encoding improves upon this simple seasonal reference.

The broader early-prediction literature similarly emphasizes that practical warning performance cannot be summarized by classification metrics alone. TEASER evaluates the trade-off between correctness and earliness \citep{schafer2020teaser}, while temporally structured event-prediction methods emphasize alignment between model output and discrete future outcomes \citep{yeche2023temporal}. Methodological studies have also shown that early-classification results can be misleading when temporal structure and alert timing are not handled explicitly \citep{wu2021early}.

Interpretability is equally important when predictions are based on environmental and calendar variables. Feature-attribution methods such as SHAP can identify which variables a model uses, but they do not establish biological causality. High importance for humidity, rainfall, or temperature may reflect direct environmental association, correlation with other predictors, or indirect encoding of seasonal position. Likewise, high importance for day-of-year variables can reflect pathogen seasonality, phenological development, recurring management schedules, or a combination of these factors.

In the present study, global SHAP analysis is used to describe how a representative fitted model distributes attribution across calendar and meteorological variables and across historical lags. Such attributions characterize model usage conditional on the included and correlated predictors; they do not establish biological causality or independent predictive contribution. The SHAP findings are therefore interpreted together with the calendar-only, environmental-only, and combined input experiments.

\subsection{Positioning of This Work}

This study is positioned as a formulation- and evaluation-oriented investigation of short-horizon vineyard risk warning rather than as the introduction of a new predictive architecture. Using treatment-associated mildew annotations, it examines whether upcoming risk episodes are predicted from recent environmental observations, recurring seasonal timing, or a combination of both. A monthly climatological baseline and learned calendar-only representations quantify the predictive value of recurring seasonal structure. The event-onset target is contrasted with conventional daily-status classification to determine whether favorable point-wise performance corresponds to useful warning of new episodes.

The contribution therefore lies in the joint analysis of target provenance, seasonal confounding, event construction, temporal sensitivity, conventional daily classification, and operational alert evaluation. Because the available annotations are management-derived and the downy- and powdery-mildew labels exhibit almost complete temporal agreement, the task is interpreted as forecasting an operational any-mildew risk proxy rather than direct biological infection onset. The findings are intended to clarify how prediction formulation and evaluation choices affect conclusions drawn from limited, proxy-labelled agricultural time series.


\section{Methodology}

\subsection{Overview of the Analysis Pipeline}

\begin{figure*}[!t]
    \centering
    \includegraphics[width=\textwidth]{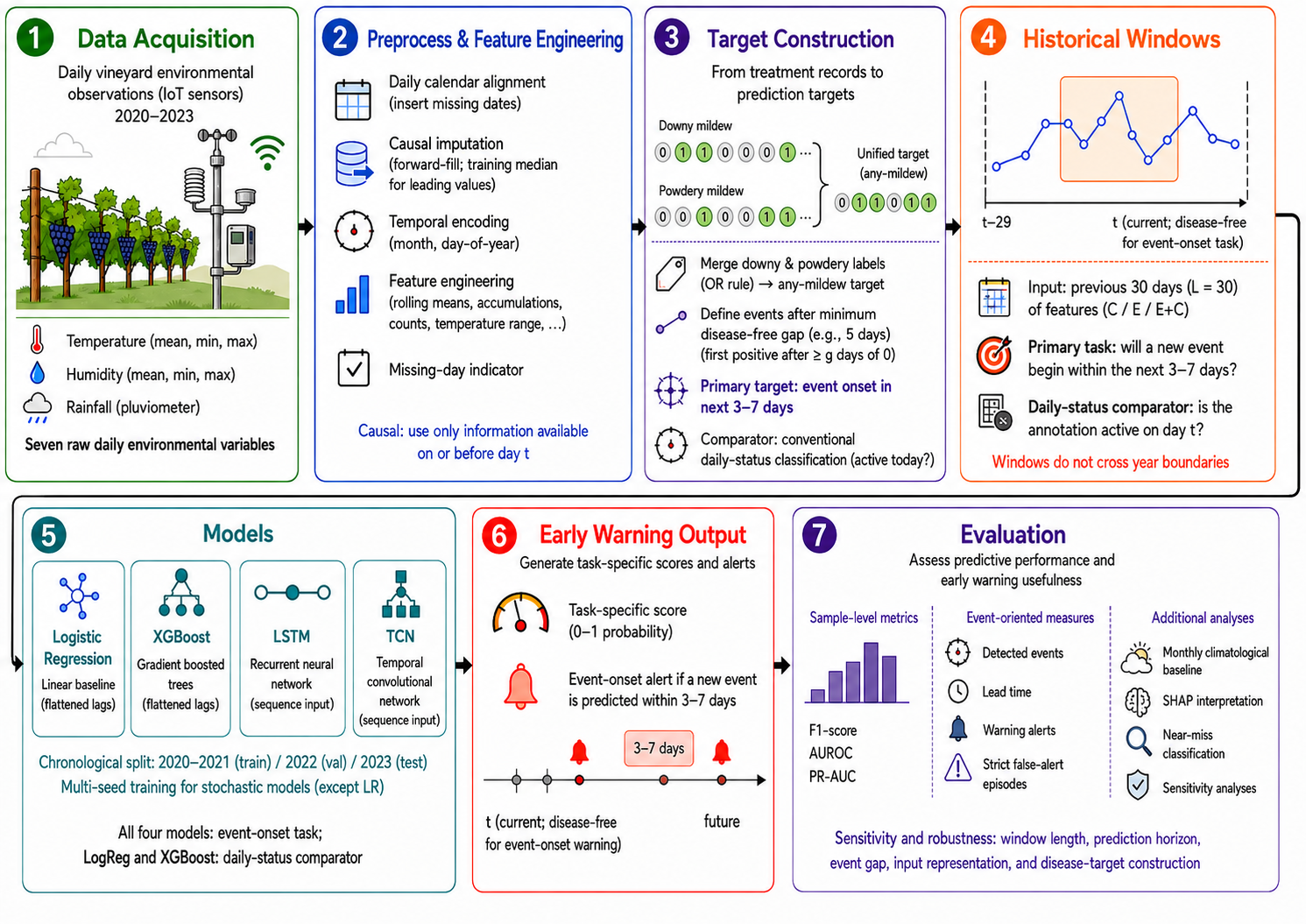}
    \caption{Overview of the analysis pipeline. The source yearly records are aligned to complete daily calendars and processed using causal environmental imputation, cyclic temporal encoding, and agro-meteorological feature construction. Downy- and powdery-mildew annotations are audited and combined into an operational any-mildew target, from which gap-separated risk events are derived. Calendar-only (C), environmental-only (E), and combined (E+C) historical representations are evaluated with logistic regression, XGBoost, LSTM, and TCN models. The event-onset formulation is compared with conventional daily-status classification. Evaluation combines sample-level metrics, event-oriented warning measures, sensitivity analysis, and model-interpretation procedures.}
    \label{fig:system_pipeline}
\end{figure*}

Figure~\ref{fig:system_pipeline} summarizes the analysis pipeline. The original yearly vineyard records are first aligned to complete daily calendars, after which environmental variables are cleaned and causal agro-meteorological descriptors are constructed. The original downy- and powdery-mildew annotations are audited, combined into an operational any-mildew target, and transformed into event indicators using a minimum disease-free gap. Historical windows are then generated under three input configurations: calendar-only (C), environmental-only (E), and combined calendar--environmental representations (E+C).

The primary task predicts whether a new treatment-associated risk event will begin within a future interval of 3--7 days. This event-onset formulation is compared with conventional daily-status classification, which determines whether the treatment-associated annotation is active on an individual day. Logistic regression, XGBoost, Long Short-Term Memory networks, and Temporal Convolutional Networks are evaluated for the primary event-onset task, while logistic regression and XGBoost provide matched learned comparators for the daily-status task. Predictions are assessed using sample-level metrics and an event-oriented evaluation framework. Sensitivity analyses examine the influence of the historical window, prediction horizon, event-separation gap, and disease-target construction, while SHAP analysis describes attribution across seasonal and environmental feature groups and historical lags.

\subsection{Dataset and Study Setting}

This study uses the vineyard agro-meteorological dataset introduced by Arvanitis et al.~\citep{arvanitis2025utilizing}. The dataset contains daily observations from a single vineyard monitoring setting covering 2020--2023. Seven environmental variables are available:

\begin{itemize}
    \item mean, minimum, and maximum relative humidity;
    \item mean, minimum, and maximum air temperature;
    \item daily pluviometer measurements.
\end{itemize}

The environmental observations are accompanied by binary downy- and powdery-mildew annotations derived from historical fungicide-treatment records. These annotations should therefore be interpreted as management-oriented disease-risk proxies rather than as direct pathological measurements of infection, symptom onset, or disease severity. Treatment timing may reflect weather-related risk, but may also be influenced by preventive schedules, phenological development, grower practices, previous field observations, and operational constraints.

\begin{figure*}[!t]
    \centering
    \includegraphics[width=\linewidth]
    {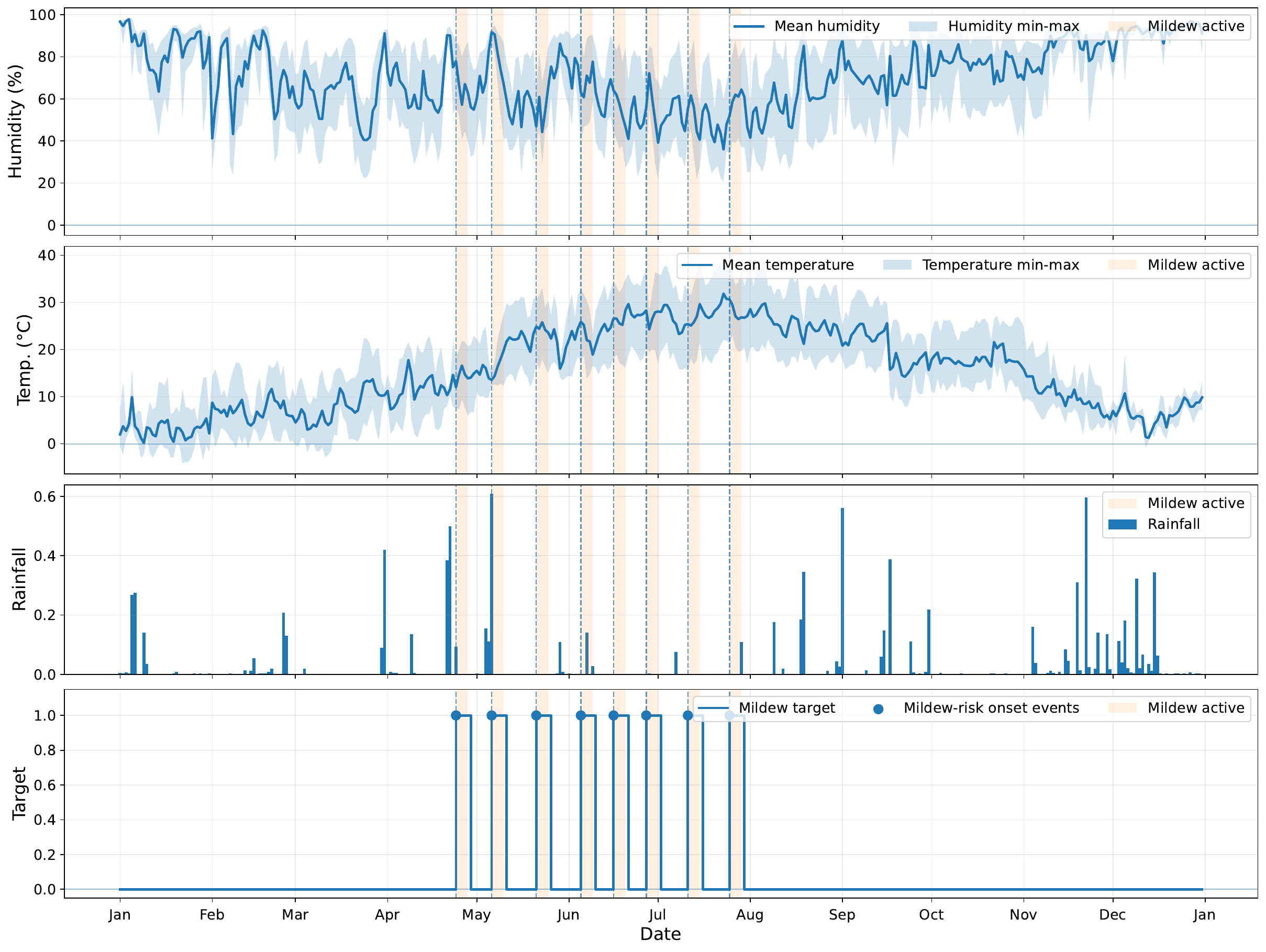}
    \caption{Example of the vineyard environmental time series and derived risk annotations for the 2022 season. The upper panels show daily humidity, temperature, and rainfall measurements. The lower panel shows the unified any-mildew target obtained from the original treatment-associated downy- and powdery-mildew annotations. Shaded intervals indicate positive daily risk periods, while vertical markers identify gap-separated event dates derived using the reference disease-free gap $g=5$. The figure illustrates the distinction between persistent daily status and the discrete transitions used in the event-onset forecasting task.}
    \label{fig:dataset_timeseries}
\end{figure*}

Figure~\ref{fig:dataset_timeseries} illustrates the complete 2022 annual time series. It shows the daily humidity, temperature, and rainfall observations together with the unified treatment-associated mildew-risk target and the event indicators obtained using the minimum disease-free gap rule. The figure highlights both the persistence of the daily binary annotations and the smaller number of discrete transitions retained by the event-based formulation.

The source data are supplied as four yearly tables. Across the observed date ranges, nine calendar dates are absent from the original files. The source annotations are also highly similar across the two disease columns: downy- and powdery-mildew labels agree on 98.9\% of observed days and are identical throughout the validation and test years. Table~\ref{tab:label_composition} summarizes their composition.

\begin{table*}[t]
\centering
\caption{Yearly composition of the original disease-related annotations. Event counts use the reference disease-free gap $g=5$.}
\label{tab:label_composition}
\setlength{\tabcolsep}{4.5pt}
\begin{tabular}{lccccccc}
\toprule
Year &
\shortstack{Observed\\days} &
\shortstack{Missing\\days} &
\shortstack{Downy\\positive days} &
\shortstack{Powdery\\positive days} &
\shortstack{Downy\\events} &
\shortstack{Powdery\\events} &
\shortstack{Unified\\events} \\
\midrule
2020 & 306 & 3 & 44 & 54 & 7 & 8 & 8 \\
2021 & 362 & 3 & 45 & 40 & 8 & 8 & 8 \\
2022 & 364 & 1 & 40 & 40 & 8 & 8 & 8 \\
2023 & 363 & 2 & 60 & 60 & 8 & 8 & 8 \\
\bottomrule
\end{tabular}
\end{table*}

Although downy and powdery mildew have distinct and partly contrasting epidemiological drivers, the near-equivalence of their available annotations provides very limited pathogen-specific variation. They are therefore combined using a logical OR operation to define an operational any-mildew target. This choice does not imply biological or epidemiological equivalence. Separate downy, powdery, OR, and AND constructions are evaluated as robustness checks, but the data do not support pathogen-specific mechanistic conclusions.

The study is conducted as a single-site temporal case study. Data from 2020--2021 are used for training, 2022 is used for validation and model selection, and 2023 is reserved for final testing. This chronological protocol approximates the intended deployment setting, in which models trained on previous seasons are applied to a future unseen season. It does not support claims of cross-site or cross-region generalization.

\subsection{Daily Calendar Alignment and Missing Data}

Each yearly table is sorted by date, duplicate dates are resolved by retaining the final record, and the data are reindexed to a complete daily calendar within the observed date range of that year. A binary missing-day indicator is added to distinguish inserted dates from originally observed records.

Environmental measurements on inserted dates are imputed using a causal procedure. Missing values for which an earlier observation is available are forward-filled within the same yearly sequence. Leading missing values, for which no previous observation exists, are replaced using the corresponding variable median estimated exclusively from the 2020--2021 training data. The missing-day indicator is retained so that the models can distinguish observed from imputed calendar dates. No future observation from the validation or test sequence is used during imputation.

Disease labels on inserted dates are not imputed. They remain unknown because the absence of a source row cannot be interpreted as evidence of either positive or negative disease risk. Samples are excluded whenever the current-state condition, event-separation rule, or target construction depends on an unknown label. This avoids silently treating missing annotations as disease-free days.

\subsection{Target Construction}

\subsubsection{Unified Daily Risk Target}

Let $d_t,p_t\in\{0,1\}$ denote the observed downy- and powdery-mildew annotations on day $t$. When both labels are available, the operational any-mildew target is defined as

\begin{equation}
m_t = d_t \lor p_t.
\label{eq:unified_target}
\end{equation}

If either source annotation is unavailable because the date was absent from the source data, $m_t$ is treated as unknown.

The union in Equation~\ref{eq:unified_target} indicates that either treatment-associated annotation is sufficient to mark the day as positive. It does not imply that the two pathogens share identical environmental requirements.

\subsubsection{Gap-Separated Event Definition}

A new event is defined as the first positive day following at least $g$ consecutive observed disease-free days. Formally,

\begin{equation}
e_t^{(g)}
=
\mathbb{I}
\left[
m_t=1
\;\land\;
m_{t-1}=m_{t-2}=\cdots=m_{t-g}=0
\right],
\label{eq:event_definition}
\end{equation}

where $\mathbb{I}[\cdot]$ denotes the indicator function. The event indicator is considered defined only when $m_t$ and all $g$ preceding labels are observed.

The gap rule is an operational episode-separation mechanism. It prevents brief interruptions in a treatment-associated positive period from automatically being counted as independent events. It should not be interpreted as a pathogen-specific incubation period or biological latency assumption. The reference analysis uses $g=5$, with $g\in\{3,5,7\}$ evaluated in sensitivity experiments.

\subsubsection{Primary Event-Onset Target}

Let $\mathbf{x}_t\in\mathbb{R}^{d}$ denote the feature vector observed on day $t$. Given a historical window of length $L$, the input sequence is

\begin{equation}
\mathbf{X}_t
=
[\mathbf{x}_{t-L+1},\ldots,\mathbf{x}_{t}].
\label{eq:historical_window}
\end{equation}

For a future horizon $[h_0,h_1]$, the event-onset target is

\begin{equation}
y_t^{\mathrm{event}}
=
\mathbb{I}
\left[
\exists\,\tau\in
\{t+h_0,\ldots,t+h_1\}
:
e_\tau^{(g)}=1
\right].
\label{eq:event_target}
\end{equation}

Only prediction days satisfying $m_t=0$ are eligible. This formulation assumes that the current operational annotation status is available at prediction time, for example from the vineyard management record. The model does not infer the current value of $m_t$ from the environmental inputs. The current-day annotation is used exclusively to restrict the evaluation to periods in which a new episode could meaningfully be anticipated; it is not included as an input feature.

The reference configuration uses

\[
L=30,\qquad g=5,\qquad [h_0,h_1]=[3,7].
\]

The corresponding operational question is:

\begin{quote}
Given observations available up to a day on which the operational treatment-associated annotation is inactive, will a new treatment-associated risk episode begin within the next 3--7 days?
\end{quote}

This formulation targets transitions into new positive episodes rather than continuation of an already active status. It does not assume that the event date represents pathologically confirmed infection onset.

\subsubsection{Daily-Status Classification Comparator}

To compare the event-onset formulation with conventional point-wise classification, a daily-status target is defined as

\begin{equation}
y_t^{\mathrm{daily}} = m_t.
\label{eq:daily_status_target}
\end{equation}

This task asks whether the treatment-associated any-mildew annotation is active on day $t$. The model receives a historical environmental and calendar window ending on the day being classified, but the current or previous disease annotations are not included among the input features.

Daily-status classification is a contemporaneous classification task rather than an early-warning task. It is included to examine how performance on persistent daily annotations differs from performance when the target explicitly represents an upcoming transition. Logistic regression and XGBoost are used as matched daily-status comparators because both models are also included in the event-onset experiments, allowing the effect of target formulation to be examined without changing the model family.

Because daily-status classification and event-onset prediction define different target variables and eligible sample sets, their sample-level metrics are interpreted within each task rather than as directly interchangeable measures. The positive outputs of the daily-status models are additionally assessed under the common event-oriented protocol described in Section~4 to determine whether they provide warning before new episodes.

\subsection{Causal Integrity of Feature and Target Construction}

For every prediction time $t$, all model inputs are constructed exclusively from environmental and calendar observations available on or before $t$. Environmental imputation uses forward filling whenever a previous observation is available; leading missing values are initialized using training-derived variable medians. Rolling descriptors use only current and preceding observations, and calendar features are deterministic functions of the corresponding date. No environmental observation after $t$ is included in $\mathbf{X}_t$.

For the event-onset task, future labels are used only to construct the supervised target in Equation~\ref{eq:event_target}. The event indicator at a future date $\tau$ is retrospectively defined from $m_\tau$ and the $g$ labels immediately preceding $\tau$. No label beyond the specified future target interval is required for sample construction.

For daily-status classification, $m_t$ is used only as the supervised target in Equation~\ref{eq:daily_status_target}. Neither $m_t$ nor any preceding disease annotation is supplied as an input feature. The current daily label is used as an eligibility criterion only for the event-onset formulation and is never provided to the predictive models.

Normalization parameters are estimated exclusively from the 2020--2021 training data and are subsequently applied unchanged to validation and test observations. Historical windows are constructed independently within each year. Neither an input window nor an event-onset target interval is permitted to cross a year boundary. Samples are omitted if any label required to define the daily target, current eligibility status, event indicator, or future event-onset target is unknown.

\subsection{Input Representations}

Three input configurations are constructed to examine the relative contribution of seasonal timing and agro-meteorological information.

\subsubsection{Calendar-Only Representation}

The calendar-only representation, denoted by C, contains cyclic encodings of month and day of year:

\begin{equation}
\begin{aligned}
\mathrm{month}_{\sin}(t)
&=
\sin\left(
2\pi\frac{\mathrm{month}(t)}{12}
\right),\\
\mathrm{month}_{\cos}(t)
&=
\cos\left(
2\pi\frac{\mathrm{month}(t)}{12}
\right).
\end{aligned}
\label{eq:month_encoding}
\end{equation}

For day-of-year encoding,

\begin{equation}
\begin{aligned}
\mathrm{doy}_{\sin}(t)
&=
\sin\left(
2\pi\frac{\mathrm{doy}(t)}{D_t}
\right),\\
\mathrm{doy}_{\cos}(t)
&=
\cos\left(
2\pi\frac{\mathrm{doy}(t)}{D_t}
\right),
\end{aligned}
\label{eq:doy_encoding}
\end{equation}

where

\begin{equation}
D_t=
\begin{cases}
366, & \text{if the year containing }t\text{ is a leap year},\\
365, & \text{otherwise}.
\end{cases}
\end{equation}

This representation contains four daily features. It serves as an explicit seasonal baseline and measures how much of the target structure can be predicted without environmental observations.

\subsubsection{Environmental Representation}

The environmental-only representation, denoted by E, contains the seven raw daily variables and causally constructed agro-meteorological descriptors.

For each window length $w\in\{3,5,7\}$, cumulative rainfall is calculated as

\begin{equation}
R_t^{(w)}
=
\sum_{j=0}^{w-1}
\mathrm{pluvm}_{t-j},
\end{equation}

while rolling humidity and temperature means are

\begin{equation}
H_t^{(w)}
=
\frac{1}{w}
\sum_{j=0}^{w-1}
\mathrm{humidity}_{t-j},
\end{equation}

\begin{equation}
T_t^{(w)}
=
\frac{1}{w}
\sum_{j=0}^{w-1}
\mathrm{temperature}_{t-j}.
\end{equation}

The daily temperature range is defined as

\begin{equation}
\Delta T_t
=
T_t^{\max}-T_t^{\min}.
\end{equation}

Counts of humid and rainy days are constructed for $w\in\{5,7\}$:

\begin{equation}
N_{H,t}^{(w)}
=
\sum_{j=0}^{w-1}
\mathbb{I}
[
\mathrm{humidity}_{t-j}>85
],
\end{equation}

\begin{equation}
N_{R,t}^{(w)}
=
\sum_{j=0}^{w-1}
\mathbb{I}
[
\mathrm{pluvm}_{t-j}>0
].
\end{equation}

The environmental configuration therefore contains:

\begin{itemize}
    \item seven raw humidity, temperature, and rainfall variables;
    \item three rainfall-accumulation descriptors;
    \item three rolling-humidity means;
    \item three rolling-temperature means;
    \item one temperature-range descriptor;
    \item two humid-day counts;
    \item two rainy-day counts;
    \item one missing-day indicator.
\end{itemize}

This results in 22 daily environmental features.

\subsubsection{Combined Representation}

The combined representation, denoted by E+C, contains all 22 environmental variables together with the four cyclic calendar variables, resulting in 26 features per day.

Comparing C, E, and E+C allows the analysis to distinguish explicit seasonal predictability from predictability associated with recent environmental observations. This comparison does not fully eliminate seasonality from the E configuration because temperature and other meteorological variables themselves vary seasonally. It nevertheless provides a direct test of whether explicit calendar information contributes beyond the measured environmental history.

\subsection{Normalization and Historical Window Construction}

All features included in the selected C, E, or E+C representation are standardized using

\begin{equation}
\widetilde{x}
=
\frac{x-\mu_{\mathrm{train}}}
{\sigma_{\mathrm{train}}},
\end{equation}

where $\mu_{\mathrm{train}}$ and $\sigma_{\mathrm{train}}$ are estimated using daily observations from the 2020--2021 training years only. The same parameters are applied to the 2022 validation and 2023 test years.

Fixed-length windows are extracted independently for each year. The reference event-onset analysis uses $L=30$, while $L\in\{7,14,30\}$ is evaluated in the historical-window sensitivity analysis. The daily-status comparator uses the same $L=30$ historical-window length as the reference event-onset configuration.

Before model input construction, task-specific sample eligibility rules are applied to exclude windows with incomplete historical coverage, unknown required labels, invalid current-state conditions, or target intervals extending beyond the corresponding yearly sequence. A candidate prediction window is retained only when:

\begin{enumerate}
    \item the complete historical sequence is available within the same year;
    \item the target label for the corresponding formulation is observed;
    \item the current day is disease-free when required by the event-onset formulation;
    \item every label needed to construct the event indicator and event-onset target is observed;
    \item the complete future target horizon remains within the same year for the event-onset task.
\end{enumerate}

For logistic regression and XGBoost, the $L\times d$ sequence is flattened into a lagged tabular vector. For LSTM and TCN, it is retained as an ordered multivariate sequence.

\subsection{Predictive Models}

Four established model families are compared. Their purpose is to provide complementary modelling perspectives rather than to establish a comprehensive architecture benchmark.

\subsubsection{Logistic Regression}

Logistic regression provides an interpretable linear baseline~\cite{hosmer2013applied}. Each historical window is flattened into a vector, and the probability of the corresponding binary target is modelled as

\begin{equation}
P(y_t=1\mid\mathbf{X}_t)
=
\sigma
\left(
\beta_0+\boldsymbol{\beta}^{\top}
\operatorname{vec}(\mathbf{X}_t)
\right),
\end{equation}

where $\sigma(\cdot)$ is the logistic function. Class-balanced weighting is used during fitting. Because the preprocessing, solver, and optimization settings are deterministic, logistic regression is evaluated once rather than across random seeds.

\subsubsection{XGBoost}

XGBoost is included as a nonlinear tree-based model for flattened historical windows \citep{chen2016xgboost}. Temporal structure is represented implicitly through lagged variables. Gradient-boosted trees can model nonlinear thresholds and interactions among humidity, rainfall, temperature, calendar, and lagged descriptors while remaining suitable for relatively small tabular datasets.

The reference configuration uses 400 estimators, learning rate $0.05$, maximum tree depth 4, row subsampling of $0.9$, column subsampling of $0.9$, and histogram-based tree construction. The positive class is weighted according to the class ratio in the corresponding training data.

\subsubsection{Long Short-Term Memory Network}

The LSTM model processes each $L\times d$ input as an ordered sequence \citep{siami2019performance}. It contains two unidirectional recurrent layers with hidden dimension 64 and dropout 0.2. The final hidden state is passed to a linear classification head that produces one event-onset logit.

The LSTM preserves the ordering of observations and can represent dependencies across the historical sequence. Its inclusion does not assume that the available labels encode biological accumulation mechanisms; rather, it tests whether sequential representation improves predictive performance relative to linear and tree-based alternatives.

\subsubsection{Temporal Convolutional Network}

The TCN provides a convolutional sequential alternative \citep{hewage2020temporal}. It contains three residual temporal-convolution blocks with hidden dimension 64, kernel size 3, dropout 0.2, and dilation factors 1, 2, and 4. Causal padding ensures that the representation at each time point depends only on current and previous observations. The final temporal representation is passed to a linear classification head that produces one event-onset logit.

The dilated receptive field enables the network to combine short- and medium-range temporal information without recurrent processing.

\subsubsection{Architectural Comparison}

\begin{figure*}[!t]
    \centering
    \includegraphics[width=\textwidth]{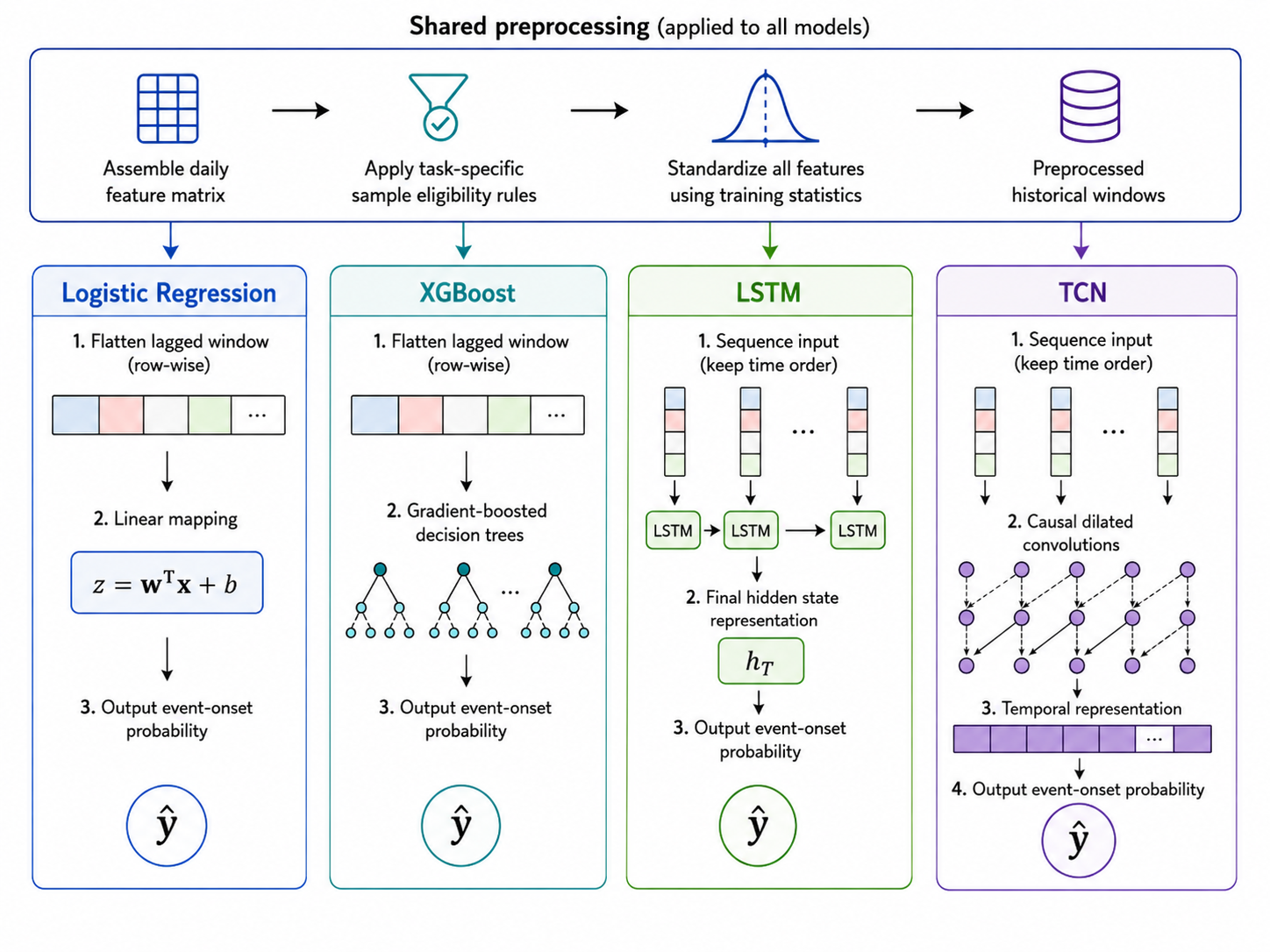}
    \caption{Compared model representations for the primary event-onset task. Logistic regression and XGBoost receive flattened lagged versions of the historical window, whereas LSTM and TCN preserve the ordered multivariate sequence. Logistic regression provides a linear baseline, XGBoost models nonlinear tabular interactions, LSTM processes the sequence recurrently, and TCN extracts temporal features using causal dilated convolutions. Logistic regression and XGBoost are additionally used as matched models for conventional daily-status classification.}
    \label{fig:model_architectures}
\end{figure*}

Logistic regression and XGBoost treat each historical window as a flattened lagged vector. Logistic regression models additive linear contributions, whereas XGBoost captures nonlinear thresholds and interactions. LSTM and TCN retain the ordered multivariate sequence: LSTM represents it through recurrent hidden states, while TCN uses hierarchical causal convolutions. The compared model representations are illustrated in Figure~\ref{fig:model_architectures}.

All four model families are compared for the primary event-onset task. Logistic regression and XGBoost are additionally used for conventional daily-status classification, providing a matched comparison between target formulations without introducing a different model family.

The comparison is intended to determine whether conclusions are stable across modelling families. It is not assumed that the most complex model should be superior, particularly because the effective sample size is limited by the number of seasons and independent events rather than by the number of overlapping windows.

The moderate computational requirements of these architectures may also facilitate future deployment-oriented evaluation on agricultural IoT platforms, although latency, memory consumption, energy use, and embedded-device performance are outside the scope of the present study.

\section{Experimental Design and Setup}

\subsection{Chronological Data Split}

All experiments follow a strictly chronological split by year. Data from 2020 and 2021 are used for model training, data from 2022 are used for validation, hyperparameter selection, and decision-threshold selection, and data from 2023 are reserved for final testing.

This protocol prevents information from future seasons from entering model development and provides a more realistic assessment of temporal generalization than random sample splitting. It also reflects the intended deployment setting, in which a model trained on historical seasons is applied to a previously unseen future season.

Historical windows and prediction targets are constructed independently within each year. Neither the input sequence nor the future target interval is allowed to cross a year boundary. Samples are excluded if any disease-related label required to determine the current status, event definition, or corresponding target is unknown.

\subsection{Training Strategy}

The reference event-onset formulation uses a historical window of $L=30$ days, a minimum disease-free gap of $g=5$ days, and a future prediction horizon of 3--7 days. Alternative values are evaluated separately through sensitivity experiments. The selected model families represent complementary learning mechanisms. Logistic regression provides an interpretable linear baseline, XGBoost captures nonlinear interactions and threshold-like relationships among lagged variables, and LSTM and TCN models preserve temporal order and can represent cumulative or sequential dependencies. Comparing these architectures helps determine whether the relative contribution of calendar timing and environmental observations is consistent across different modelling assumptions rather than being specific to a single learning mechanism.

Logistic regression and XGBoost receive flattened lagged representations of the historical window, whereas LSTM and TCN models receive the corresponding ordered multivariate sequences. All features included in the selected C, E, or E+C representation are standardized using parameters estimated exclusively from the 2020--2021 training data and applied unchanged to the validation and test sets.

Logistic regression is trained using class-balanced weighting and the liblinear solver, with a maximum of 5000 optimization iterations. The regularization strength is fixed at $C=1$. Because the preprocessing, solver, and optimization procedure are deterministic, logistic regression is evaluated once for each configuration.

For XGBoost, the principal hyperparameters are 400 estimators, a learning rate of 0.05, maximum tree depth of 4, row subsampling of 0.9, column subsampling of 0.9, and histogram-based tree construction. Binary logistic classification is used as the learning objective, with logarithmic loss as the evaluation metric. Training is performed using one processing thread to improve reproducibility. The positive class is weighted according to the class ratio in the corresponding training split. XGBoost is evaluated across multiple random seeds because row and feature subsampling introduce stochastic variation.

The LSTM and TCN models are trained using weighted binary cross-entropy. For each experiment, the positive-class weight is recomputed from the corresponding training samples as

\begin{equation}
w_{+}
=
\frac{N_{-}}{N_{+}}.
\end{equation}

For the reference event-onset configuration, the training set contains 75 positive and 377 negative windows, giving

\begin{equation}
w_{+}
=
\frac{377}{75}
=
5.0267.
\end{equation}

The same weighting principle is applied separately to alternative historical windows, prediction horizons, event gaps, disease-target constructions, and daily-status classification because these experiments can produce different positive and negative sample counts.

Neural models are trained using AdamW with learning rate $10^{-3}$, weight decay $10^{-4}$, and batch size 32. The gradient norm is clipped to a maximum value of 1.0. Early stopping is based on validation loss, with a maximum of 50 epochs and patience of 10 epochs. The LSTM uses two unidirectional recurrent layers, hidden dimension 64, and dropout 0.2. The TCN uses three residual temporal-convolutional levels, hidden dimension 64, kernel size 3, dilation factors 1, 2, and 4, and dropout 0.2.

The stochastic models are trained using five random seeds, $\{0,1,2,3,4\}$. The seed controls model initialization and all supported stochastic sampling operations. Reported results are expressed as mean and standard deviation across these five runs. Random-seed variation is used to assess optimization stability and should not be interpreted as resolving the statistical uncertainty associated with the small number of independent test events.

The selected model configurations were chosen as compact, commonly used settings appropriate for the limited size of the dataset. Preliminary model-development experiments were conducted using the 2022 validation year, while the 2023 test year remained untouched. The principal architectural settings, including the number of estimators, hidden dimension, number of layers, kernel size, and dropout rate, were fixed before final test evaluation. For the neural models, early stopping was performed using validation loss. No model configuration or hyperparameter was selected using the 2023 test data.

For each trained model, the operating threshold is selected exclusively on the 2022 validation set. Candidate thresholds are evaluated over the grid

\begin{equation}
\mathcal{T}
=
\{0.05,0.10,\ldots,0.95\}.
\end{equation}

The threshold maximizing validation F1-score is selected independently for each model, random seed, input representation, and target formulation. The selected threshold is then applied unchanged to the 2023 test set.

Transformer-based models are not included because the dataset contains only four annual cycles and a small number of independent treatment-associated events. Although overlapping windows increase the nominal number of samples, they do not create an equivalent number of independent training sequences. The study therefore focuses on lower-capacity linear, tree-based, recurrent, and convolutional models that are more appropriate for this limited-data setting.

\subsection{Baseline and Comparator Configurations}

Several baseline and comparator configurations are used to determine whether predictive performance arises from recent environmental conditions, recurring seasonal timing, or the chosen target formulation. Because the annotations are derived from fungicide-treatment records, explicit seasonal baselines are essential for determining how much performance can be obtained from recurring calendar structure without relying on recent environmental observations.

Three feature representations are compared:

\begin{itemize}
    \item \textbf{C}: cyclic month and day-of-year features only;
    \item \textbf{E}: raw and engineered environmental variables only;
    \item \textbf{E+C}: combined environmental and calendar features.
\end{itemize}

Calendar-only logistic regression provides a learned seasonal baseline. In addition, a non-parametric monthly climatological baseline estimates event-onset probability from the empirical positive rate among eligible event-onset samples from the same calendar month in the 2020--2021 training years.

Let $\mathcal{T}_{\mathrm{train}}^{\mathrm{event}}$ denote the set of eligible event-onset prediction days in the 2020--2021 training years. A day $s$ belongs to this set only if its complete historical window is
available within the corresponding year, the current annotation is observed and satisfies $m_s=0$, all labels required to define the future event-onset target are observed, and the complete prediction horizon remains within the same year. For a test or validation prediction day $t$, the monthly climatological probability is defined as

\begin{equation}
\widehat{p}^{\,\mathrm{clim}}_t
=
\frac{
\displaystyle
\sum_{s\in\mathcal{T}_{\mathrm{train}}^{\mathrm{event}}}
\mathbb{I}
\left[
\mathrm{month}(s)=\mathrm{month}(t)
\right]
y_s^{\mathrm{event}}
}{
\displaystyle
\sum_{s\in\mathcal{T}_{\mathrm{train}}^{\mathrm{event}}}
\mathbb{I}
\left[
\mathrm{month}(s)=\mathrm{month}(t)
\right]
}.
\label{eq:monthly_climatology}
\end{equation}

The numerator counts eligible training samples from the corresponding calendar month whose event-onset target is positive, whereas the denominator counts all eligible event-onset samples from that month. The baseline therefore estimates the monthly probability of an upcoming event within the 3--7-day horizon conditional on the same eligibility rules used for the learned event-onset models.

The monthly climatological baseline uses no environmental observations and no fitted machine-learning model. Its operating threshold is selected on the 2022 validation year using the same threshold grid and F1-based procedure applied to the learned models, and the selected threshold is then applied unchanged to the 2023 test year. The baseline follows the same disease-free eligibility rule, event definition, prediction horizon, and event-oriented evaluation protocol as the primary event-onset models.

The event-onset formulation is also compared with conventional daily-status classification. The daily-status comparator uses the combined E+C representation and the same $L=30$ historical window, chronological split, preprocessing procedure, and validation-based threshold selection. Logistic regression and XGBoost are used for this comparison because both are also evaluated on the event-onset task, allowing the influence of target formulation to be examined without changing the model family.

Daily-status classification and event-onset prediction define different target variables and eligible sample sets. Their F1, AUROC, and PR-AUC values are therefore interpreted within their respective tasks rather than as directly interchangeable estimates of performance. To provide a common operational comparison, positive outputs from the daily-status models are additionally evaluated using the event-oriented warning protocol described below.

\subsection{Sensitivity and Target-Robustness Analyses}

The reference values $L=30$, $g=5$, and a 3--7-day horizon are treated as operational design choices rather than universally optimal agronomic constants. Their influence is assessed through one-factor-at-a-time sensitivity experiments. Historical-window sensitivity is evaluated using

\begin{equation}
L\in\{7,14,30\}.
\end{equation}

Event-gap sensitivity is evaluated using

\begin{equation}
g\in\{3,5,7\}.
\end{equation}

Prediction-horizon sensitivity is evaluated using

\begin{equation}
[h_0,h_1]
\in
\{[1,3],[3,7],[5,10]\}.
\end{equation}

When one parameter is varied, the remaining parameters are held at their reference values. This design avoids an exhaustive retrospective search over all parameter combinations and allows the effect of each choice on sample counts, event counts, discrimination, lead time, and alert burden to be interpreted separately.

The robustness of the merged disease target is also examined using four alternative constructions:

\begin{itemize}
    \item unified OR target, $d_t\lor p_t$;
    \item consensus AND target, $d_t\land p_t$;
    \item downy-mildew target;
    \item powdery-mildew target.
\end{itemize}

These analyses determine whether the small number of discordant source labels materially affects the conclusions. Because the downy- and powdery-mildew annotations are nearly identical, especially in the validation and test years, the separate-target experiments are treated as robustness checks rather than reliable pathogen-specific forecasting analyses.

The sensitivity analyses were conducted after fixing the reference configuration and are interpreted as post hoc robustness checks. Their test-year results were not used to retune the primary models, redefine the principal target, or select the operating configuration reported in the main comparison.

\subsection{Model-Interpretation Analysis}

To examine how the representative fitted model used seasonal and environmental variables, global TreeSHAP analysis was performed for the XGBoost model trained with the full combined E+C representation using random seed 0~\citep{lundberg2017unified}. The analysis uses the same reference configuration as the main experiment: $L=30$, $g=5$, and a prediction horizon of 3--7 days. SHAP values are calculated for the flattened historical windows of the 2023 test set. This produces one attribution value for every feature--lag combination in each test sample.

To summarize global model usage, the absolute SHAP values are first averaged across all test samples. Because each daily variable appears at multiple positions within the historical window, the resulting attributions are then aggregated in two complementary ways:

\begin{itemize}
    \item by feature group: calendar, humidity, rainfall, temperature, and missingness;
    \item by lag interval: 0--2, 3--6, 7--13, and 14--29 days before the prediction date.
\end{itemize}

For presentation, cumulative attribution shares were calculated by summing the mean absolute SHAP values within each feature group or lag interval and dividing by the total absolute attribution. Because the feature groups contain different numbers of feature--lag columns and the lag intervals contain different numbers of historical positions, these cumulative percentages are influenced by group size. We therefore also calculated a size-normalized attribution index. For feature groups, the cumulative attribution share was divided by the corresponding share of feature--lag columns. For lag intervals, the cumulative attribution share was divided by the corresponding share of the 30 historical positions. An index of 1 denotes average attribution, values above 1 denote above-average attribution per column or lag, and values below 1 denote below-average attribution.

The SHAP analysis is descriptive and is based on one representative XGBoost run. It does not quantify attribution variability across random seeds and is not used for model selection or performance estimation. Moreover, SHAP values characterize model usage conditional on the included and potentially correlated predictors; they should not be interpreted as estimates of biological causality or as evidence that a feature group provides independent predictive value.

\subsection{Evaluation Protocol}

Evaluation combines sample-level classification metrics with event-oriented early-warning measures.

\subsubsection{Sample-Level Metrics}

The sample-level evaluation reports F1-score, Area Under the Receiver Operating Characteristic Curve (AUROC), and area under the precision--recall curve (PR-AUC). F1-score evaluates threshold-dependent classification performance, whereas AUROC and PR-AUC assess ranking performance across decision thresholds. PR-AUC is particularly informative in the present imbalanced setting.

These metrics assess discrimination at the prediction-window level. They do not, by themselves, indicate whether a model provides timely warning of a new risk episode.

\subsubsection{Event-Oriented Evaluation}

For the primary event-onset formulation, all eligible prediction days are already required to satisfy $m_t=0$. For the common event-oriented evaluation of conventional daily-status models, only positive outputs issued while the current annotation is disease-free are retained as candidate warning alerts. Daily-status predictions issued while $m_t=1$ contribute to the native daily-status sample-level metrics but are excluded from the warning stream because the annotated episode is already active.

For a daily-status classifier, a positive output on a day satisfying $m_t=0$ is a false positive with respect to its native contemporaneous target. In the common event-oriented analysis, such outputs are reinterpreted post hoc as candidate warnings and are counted as valid only when they fall within a future event-warning interval. This analysis evaluates the temporal usefulness of the outputs but does not convert the daily-status classifier into a prospectively trained forecasting model.

Let $d_i$ denote the date of the $i$th true event. For the reference 3--7-day horizon, its valid warning interval is

\begin{equation}
W_i=[d_i-7,d_i-3].
\end{equation}

An event is considered detected if at least one positive daily alert occurs within $W_i$. Event recall is defined as

\begin{equation}
\mathrm{EventRecall}
=
\frac{\text{number of detected events}}
{\text{total number of events}}.
\end{equation}

Because the number of test events is small, results are principally presented as raw detected-event counts together with the total number of reference events.

If multiple valid alerts precede the same event, lead time is calculated from the earliest valid alert:

\begin{equation}
\mathrm{LeadTime}_i
=
d_i-
\min
\left\{
a:a\in W_i
\right\}.
\end{equation}

Mean lead time is calculated over detected events only. Daily alerts are grouped into alert episodes. An episode is defined as a maximal sequence of alerts occurring on consecutive calendar days. Therefore, two alerts belong to the same episode when the difference between their dates is no greater than one day.

To distinguish slightly mistimed alerts from unrelated warnings, a two-day tolerance is defined immediately outside each valid interval:

\begin{equation}
N_i
=
[d_i-9,d_i-8]
\cup
[d_i-2,d_i-1].
\end{equation}

This tolerance is particularly relevant because the target dates originate from management records rather than direct observations of biological infection onset. Treatment timing may not coincide exactly with the underlying period of elevated environmental risk and may also be affected by management and operational considerations. Alerts occurring close to, but outside, the predefined warning interval are therefore distinguished from alerts that are temporally unrelated to any annotated event. This classification does not assume that near-miss alerts are biologically correct; it provides a more transparent description of their temporal proximity to the proxy event.

An alert is classified as:

\begin{itemize}
    \item \textbf{valid} if it occurs within any $W_i$;
    \item \textbf{near miss} if it falls outside all valid intervals but within at least one $N_i$;
    \item \textbf{strict false alert} if it falls outside all valid and near-miss intervals.
\end{itemize}

An episode containing at least one valid alert is treated as a successful episode. If it contains no valid alert but includes at least one near-miss alert, it is classified as a near-miss episode. All remaining episodes are classified as strict false-alert episodes.

Near-miss alerts and episodes are used to separate slightly mistimed warnings from the strict false-alert burden. Because the principal results focus on detected events, total alert days, and strict false-alert episodes, near-miss counts are not reported as separate performance metrics.

The principal event-oriented results report the number of detected events, total alert days, and strict false-alert episodes. Per-event lead times are additionally presented in Figure~\ref{fig:per_event_matrix}. Raw detected-event counts are shown alongside the total number of reference events because the 2023 test set contains only eight events.

The difference between 7 of 8 and 6 of 8 detected events is interpreted descriptively. The test set contains only eight reference events, reflecting the limited number of independent seasonal episodes available in the dataset. Consequently, a one-event difference does not establish statistical superiority of one model and is considered together with optimization variability, warning lead time, alert burden, and false-alert behavior.

\section{Results and Discussion}

\subsection{Seasonal Baseline and Input Representations}

Because the eight reference events in the 2023 test year occur within the vineyard growing season and follow recurring temporal patterns similar to those observed in the training years, we first evaluated a monthly climatological baseline. Table~\ref{tab:monthly_climatology} reports its performance on the 2023 test year. The baseline estimates event-onset probability from the empirical positive rate for the corresponding calendar month in the 2020--2021 training years and does not use environmental observations or a fitted machine-learning model.

\begin{table*}[t]
\centering
\caption{Performance of the monthly climatological baseline on the 2023 test year. The threshold was selected on the 2022 validation year. Detected events are reported out of eight reference events.}
\label{tab:monthly_climatology}

\small
\setlength{\tabcolsep}{5pt}
\renewcommand{\arraystretch}{1.08}

\begin{tabular}{lccccc}
\toprule
Baseline &
F1 &
AUROC &
PR-AUC &
\makecell{Detected\\events} &
Alerts \\
\midrule
Monthly climatology
& 0.507
& 0.830
& 0.401
& 6/8
& 42 \\
\bottomrule
\end{tabular}
\end{table*}

As shown in Table~\ref{tab:monthly_climatology}, the monthly climatological baseline achieved an F1-score of 0.507, an AUROC of 0.830, and a PR-AUC of 0.401. It detected 6 of the 8 test events while generating 42 warning alerts. This non-trivial performance shows that recurring seasonal timing contains substantial predictive information even without recent environmental observations.

Table~\ref{tab:main_feature_results} reports the corresponding results for calendar-only (C), environmental-only (E), and combined environmental and calendar (E+C) representations. Logistic regression is deterministic, whereas XGBoost, LSTM, and TCN results are reported as mean $\pm$ standard deviation over five random seeds. Decision thresholds were selected exclusively on the 2022 validation year and then applied unchanged to the 2023 test year.

\begin{table*}[t]
\centering
\caption{Performance on the 2023 test year for calendar-only (C), environmental-only (E), and combined (E+C) representations. Detected events are reported out of eight test events. Results for stochastic models are mean $\pm$ standard deviation over five seeds.}
\label{tab:main_feature_results}

\small
\setlength{\tabcolsep}{3.2pt}
\renewcommand{\arraystretch}{1.08}

\begin{tabular}{llcccccc}
\toprule
Input &
Model &
F1 &
AUROC &
PR-AUC &
\makecell{Detected\\events} &
Alerts &
\makecell{Strict false-alert\\episodes} \\
\midrule

C & LogReg
& 0.562
& 0.900
& 0.495
& 5/8
& 31
& 2 \\

C & XGBoost
& $0.495\pm0.016$
& $0.842\pm0.004$
& $0.385\pm0.007$
& $(6.8\pm0.4)/8$
& $53.4\pm0.9$
& $5.0\pm0.0$ \\

C & LSTM
& $0.638\pm0.021$
& $0.902\pm0.015$
& $0.503\pm0.055$
& $(7.0\pm0.0)/8$
& $48.6\pm2.7$
& $3.8\pm0.8$ \\

C & TCN
& $0.608\pm0.022$
& $0.913\pm0.007$
& $0.537\pm0.025$
& $(7.0\pm0.0)/8$
& $53.2\pm3.1$
& $4.2\pm0.4$ \\
\midrule

E & LogReg
& 0.327
& 0.638
& 0.215
& 5/8
& 71
& 15 \\

E & XGBoost
& $0.225\pm0.064$
& $0.651\pm0.015$
& $0.201\pm0.009$
& $(3.8\pm1.1)/8$
& $35.8\pm10.2$
& $7.8\pm2.9$ \\

E & LSTM
& $0.216\pm0.118$
& $0.646\pm0.054$
& $0.179\pm0.032$
& $(4.0\pm3.0)/8$
& $84.6\pm54.7$
& $5.6\pm2.1$ \\

E & TCN
& $0.254\pm0.048$
& $0.645\pm0.043$
& $0.189\pm0.033$
& $(5.0\pm1.6)/8$
& $64.0\pm11.6$
& $8.8\pm1.8$ \\
\midrule

E+C & LogReg
& 0.405
& 0.732
& 0.307
& 6/8
& 46
& 13 \\

E+C & XGBoost
& $0.555\pm0.028$
& $0.755\pm0.011$
& $0.350\pm0.016$
& $(6.0\pm0.0)/8$
& $41.2\pm1.9$
& $3.2\pm0.4$ \\

E+C & LSTM
& $0.560\pm0.011$
& $0.898\pm0.019$
& $0.498\pm0.075$
& $(6.6\pm0.9)/8$
& $52.8\pm5.0$
& $4.4\pm0.5$ \\

E+C & TCN
& $0.515\pm0.023$
& $0.880\pm0.002$
& $0.387\pm0.005$
& $(6.2\pm0.4)/8$
& $52.4\pm1.9$
& $4.8\pm0.4$ \\
\bottomrule
\end{tabular}
\end{table*}

The strongest and most stable finding is the competitiveness of the calendar-only representation. Calendar-only LSTM and TCN models both detected 7 of the 8 test events in every run. They achieved F1-scores of $0.638\pm0.021$ and $0.608\pm0.022$, respectively, while the calendar-only TCN obtained the highest AUROC and PR-AUC among all evaluated configurations, with values of $0.913\pm0.007$ and $0.537\pm0.025$.

The learned calendar-only models also exceeded the monthly climatological baseline on several metrics. However, the climatological baseline itself detected 6 of 8 events and outperformed several environmental-only configurations. The predictive value of calendar information therefore cannot be attributed solely to the capacity of the learned models. Both the non-parametric monthly baseline and the cyclic calendar representations reveal substantial recurring seasonal structure in the treatment-associated target.

Environmental-only models were substantially weaker and generally less stable. AUROC values for environmental-only XGBoost, LSTM, and TCN were approximately 0.64--0.65, while their PR-AUC values remained below 0.21. The environmental-only LSTM showed particularly high variability, detecting $(4.0\pm3.0)/8$ events and generating $84.6\pm54.7$ alerts. The environmental-only TCN detected more events on average but produced the largest strict false-alert episode burden. Thus, recent environmental history alone did not reproduce the stronger and more stable performance obtained from explicit calendar timing.

Adding environmental variables to calendar information produced model-dependent rather than consistent changes. For XGBoost, the combined representation increased F1 from $0.495\pm0.016$ to $0.555\pm0.028$, but reduced AUROC, PR-AUC, and mean event coverage. For LSTM and TCN, calendar-only input remained stronger across the principal discrimination and event-coverage measures. The combined LSTM retained relatively strong ranking performance, but its PR-AUC and event coverage were less stable than those of the calendar-only model.

No model family was uniformly superior. Calendar-only LSTM achieved the highest F1-score, calendar-only TCN achieved the highest AUROC and PR-AUC, and combined XGBoost generated the fewest strict false-alert episodes. These results represent configuration-specific trade-offs among discrimination, event coverage, alert burden, and stability rather than evidence of overall superiority of a single architecture.

\begin{figure*}[!t]
    \centering
    \includegraphics[width=\textwidth]{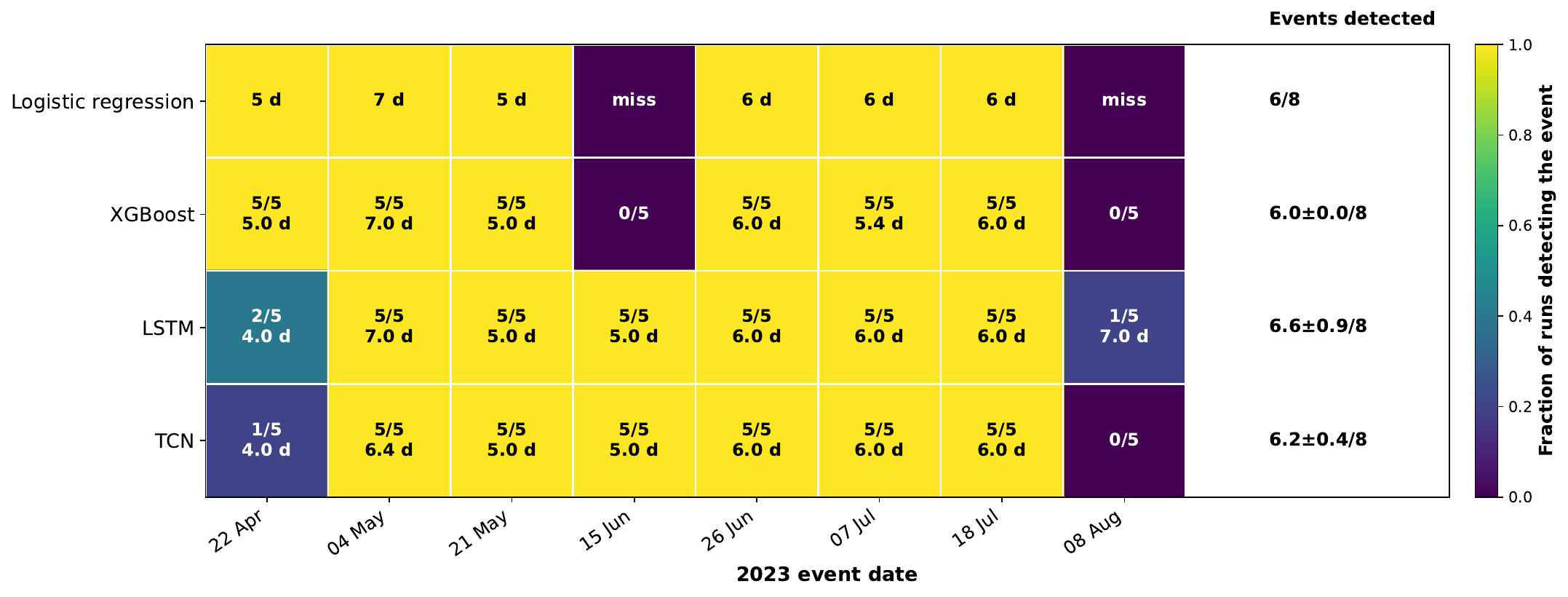}
    \caption{Per-event detection stability on the 2023 test season for the combined environmental and calendar representation ($L=30$, $g=5$, prediction horizon 3--7 days). Columns correspond to the eight true treatment-associated risk events. For XGBoost, LSTM, and TCN, each cell reports the number of runs out of five in which the event was detected and the mean lead time among successful runs. Logistic regression is deterministic and reports either its lead time or a missed-event indicator. Cell shading represents the fraction of runs detecting each event, while the rightmost values summarize the mean number of detected events across seeds. The figure uses the common E+C reference configuration and does not represent the individually best-performing input representation for each model.}
    \label{fig:per_event_matrix}
\end{figure*}

Figure~\ref{fig:per_event_matrix} provides a more detailed comparison of event-level behavior for the common E+C reference configuration. For stochastic models, each cell reports the number of random-seed runs in which the corresponding event was detected and the mean lead time among successful runs. Logistic regression is deterministic and therefore reports either one lead time or a missed-event indicator. Most events occurring from May to July were detected consistently by the sequence models. In contrast, the event on 22 April was detected in only two of five LSTM runs and one of five TCN runs, while the final event on 8 August was detected in one LSTM run and in none of the XGBoost or TCN runs. XGBoost consistently missed the events on 15 June and 8 August.

\subsection{Event-Onset Prediction Versus Daily-Status Classification}

Conventional daily-status classification and event-onset prediction represent different operational tasks. Daily-status classification determines whether the treatment-associated annotation is active on the day being classified, whereas event-onset prediction asks whether a new episode will begin within the following 3--7 days. Their native sample-level metrics are therefore interpreted within each task rather than as directly interchangeable measures.

Table~\ref{tab:daily_event_comparison} compares logistic regression and XGBoost under both formulations using the combined E+C representation and $L=30$. Panel A reports sample-level performance on each formulation's native target. Panel B evaluates positive outputs from both formulations under the same event-oriented protocol. For the daily-status models, only positive outputs issued while the current annotation was disease-free were retained as candidate warning alerts.

\begin{table*}[t]
\centering
\caption{Comparison of conventional daily-status classification and event-onset prediction on the 2023 test year using the combined E+C representation and $L=30$. Panel A reports native sample-level performance. Panel B evaluates both formulations under the common event-oriented warning protocol. XGBoost results are mean $\pm$ standard deviation over five seeds.}
\label{tab:daily_event_comparison}

\small
\setlength{\tabcolsep}{4pt}
\renewcommand{\arraystretch}{1.08}

\begin{tabular}{llccccc}
\toprule
\multicolumn{7}{c}{\textbf{Panel A: Native task performance}} \\
\midrule
Training target &
Model &
F1 &
AUROC &
PR-AUC &
\multicolumn{2}{c}{} \\
\midrule

Daily status & LogReg
& 0.444
& 0.788
& 0.403
& \multicolumn{2}{c}{} \\

Daily status & XGBoost
& $0.517\pm0.014$
& $0.804\pm0.005$
& $0.467\pm0.015$
& \multicolumn{2}{c}{} \\

Event onset & LogReg
& 0.405
& 0.732
& 0.307
& \multicolumn{2}{c}{} \\

Event onset & XGBoost
& $0.555\pm0.028$
& $0.755\pm0.011$
& $0.350\pm0.016$
& \multicolumn{2}{c}{} \\
\midrule

\multicolumn{7}{c}{\textbf{Panel B: Common event-oriented evaluation}} \\
\midrule
Training target &
Model &
\makecell{Detected\\events} &
\makecell{Mean lead\\time} &
\makecell{Warning\\alerts} &
\makecell{Strict false-alert\\episodes} &
\\
\midrule

Daily status & LogReg
& 7/8
& 5.00
& 52
& 12
& \\

Daily status & XGBoost
& $(5.0\pm0.0)/8$
& $5.80\pm0.24$
& $37.4\pm1.1$
& $3.2\pm0.4$
& \\

Event onset & LogReg
& 6/8
& 5.83
& 46
& 13
& \\

Event onset & XGBoost
& $(6.0\pm0.0)/8$
& $5.73\pm0.15$
& $41.2\pm1.9$
& $3.2\pm0.4$
& \\
\bottomrule
\end{tabular}
\end{table*}

On their native targets, daily-status logistic regression achieved higher F1, AUROC, and PR-AUC than event-onset logistic regression. Daily-status XGBoost also achieved higher AUROC and PR-AUC than event-onset XGBoost, whereas event-onset XGBoost achieved the higher F1-score. These values should not be interpreted as a direct ranking of task difficulty because the two formulations define different positive labels and eligible samples.

The common event-oriented evaluation shows that neither formulation was uniformly superior. Daily-status logistic regression detected 7 of 8 events compared with 6 of 8 for event-onset logistic regression, but it generated 52 warning alerts rather than 46. Daily-status XGBoost generated fewer alerts than event-onset XGBoost and the same mean number of strict false episodes, but it detected one fewer event. Event-onset XGBoost therefore provided higher event coverage at the cost of approximately four additional alert days.

These findings indicate that conventional daily-status outputs can sometimes occur within valid pre-event intervals, even though the model was not explicitly trained for onset warning. However, their timing and operational behavior differ from those of models trained directly on the event-onset target. The main value of the event formulation is therefore not guaranteed numerical superiority, but explicit alignment between the supervised target and the early-warning objective.

\subsection{Sensitivity to the Historical Window}

Table~\ref{tab:window_sensitivity} reports sensitivity to historical window lengths $L\in\{7,14,30\}$ using the combined E+C representation. All other settings are held at their reference values, $g=5$ and a 3--7-day prediction horizon.

\begin{table*}[t]
\centering
\caption{Historical-window sensitivity on the 2023 test year for the combined E+C representation. Detected events are reported out of eight test events. Results for stochastic models are mean $\pm$ standard deviation over five seeds.}
\label{tab:window_sensitivity}

\small
\setlength{\tabcolsep}{4.2pt}
\renewcommand{\arraystretch}{1.08}

\begin{tabular}{lcccccc}
\toprule
Model &
$L$ &
F1 &
AUROC &
PR-AUC &
\makecell{Detected\\events} &
Alerts \\
\midrule

LogReg & 7
& 0.462
& 0.890
& 0.402
& 6/8
& 45 \\

LogReg & 14
& 0.364
& 0.859
& 0.406
& 4/8
& 22 \\

LogReg & 30
& 0.405
& 0.732
& 0.307
& 6/8
& 46 \\
\midrule

XGBoost & 7
& $0.537\pm0.020$
& $0.857\pm0.008$
& $0.523\pm0.020$
& $(6.8\pm0.4)/8$
& $50.4\pm2.1$ \\

XGBoost & 14
& $0.571\pm0.033$
& $0.871\pm0.005$
& $0.565\pm0.032$
& $(7.0\pm0.0)/8$
& $40.0\pm2.5$ \\

XGBoost & 30
& $0.555\pm0.028$
& $0.755\pm0.011$
& $0.350\pm0.016$
& $(6.0\pm0.0)/8$
& $41.2\pm1.9$ \\
\midrule

LSTM & 7
& $0.575\pm0.046$
& $0.920\pm0.005$
& $0.592\pm0.020$
& $(6.4\pm0.5)/8$
& $42.4\pm10.0$ \\

LSTM & 14
& $0.504\pm0.015$
& $0.855\pm0.012$
& $0.434\pm0.080$
& $(6.0\pm0.0)/8$
& $56.8\pm3.7$ \\

LSTM & 30
& $0.560\pm0.011$
& $0.898\pm0.019$
& $0.498\pm0.075$
& $(6.6\pm0.9)/8$
& $52.8\pm5.0$ \\
\midrule

TCN & 7
& $0.550\pm0.012$
& $0.902\pm0.004$
& $0.426\pm0.024$
& $(6.2\pm0.4)/8$
& $50.0\pm3.2$ \\

TCN & 14
& $0.510\pm0.017$
& $0.880\pm0.008$
& $0.370\pm0.013$
& $(6.0\pm0.0)/8$
& $55.0\pm4.5$ \\

TCN & 30
& $0.515\pm0.023$
& $0.880\pm0.002$
& $0.387\pm0.005$
& $(6.2\pm0.4)/8$
& $52.4\pm1.9$ \\
\bottomrule
\end{tabular}
\end{table*}

No historical window was uniformly optimal across model families. XGBoost performed best with $L=14$, achieving its highest F1, AUROC, PR-AUC, and event coverage while generating fewer alerts than its 7-day configuration. In contrast, the 7-day window produced the strongest discrimination results for both LSTM and TCN. The LSTM with $L=7$ achieved the highest AUROC and PR-AUC in the table, although its alert count showed substantial variability across seeds.

The 30-day window was not consistently superior. It remained competitive for LSTM event coverage, detecting $(6.6\pm0.9)/8$ events, but its discrimination metrics were lower than those of the 7-day configuration. For XGBoost, increasing the historical window from 14 to 30 days caused a marked reduction in AUROC and PR-AUC. TCN performance was comparatively stable between 14 and 30 days, but both settings were weaker than the 7-day configuration according to F1 and PR-AUC.

These findings indicate that the preferred historical context depends on the model family. Shorter windows may reduce redundant or weakly relevant lagged information for the sequence models, whereas XGBoost benefited from an intermediate 14-day representation. Because E+C combines meteorological variables with repeated calendar features, this analysis does not isolate the temporal range of the environmental contribution alone. The reference value $L=30$ should therefore be interpreted as an operational configuration rather than a universally optimal historical window.

\subsection{Prediction-Horizon Sensitivity}

Table~\ref{tab:horizon_sensitivity} reports sensitivity to prediction horizons of 1--3, 3--7, and 5--10 days using the combined E+C representation. The historical window and event-separation gap are held at their reference values, $L=30$ and $g=5$.

\begin{table*}[t]
\centering
\caption{Prediction-horizon sensitivity on the 2023 test year for the combined E+C representation with $L=30$ and $g=5$. Detected events are reported out of eight test events. Results for stochastic models are mean $\pm$ standard deviation over five seeds.}
\label{tab:horizon_sensitivity}

\small
\setlength{\tabcolsep}{4.2pt}
\renewcommand{\arraystretch}{1.08}

\begin{tabular}{lcccccc}
\toprule
Model &
Horizon &
F1 &
AUROC &
PR-AUC &
\makecell{Detected\\events} &
Alerts \\
\midrule

LogReg & 1--3
& 0.300
& 0.764
& 0.350
& 3/8
& 16 \\

LogReg & 3--7
& 0.405
& 0.732
& 0.307
& 6/8
& 46 \\

LogReg & 5--10
& 0.164
& 0.586
& 0.138
& 4/8
& 50 \\
\midrule

XGBoost & 1--3
& $0.133\pm0.067$
& $0.702\pm0.013$
& $0.183\pm0.016$
& $(2.6\pm1.1)/8$
& $19.2\pm6.8$ \\

XGBoost & 3--7
& $0.555\pm0.028$
& $0.755\pm0.011$
& $0.350\pm0.016$
& $(6.0\pm0.0)/8$
& $41.2\pm1.9$ \\

XGBoost & 5--10
& $0.298\pm0.077$
& $0.653\pm0.006$
& $0.188\pm0.003$
& $(4.4\pm1.5)/8$
& $34.0\pm11.1$ \\
\midrule

LSTM & 1--3
& $0.490\pm0.077$
& $0.890\pm0.025$
& $0.425\pm0.026$
& $(5.6\pm1.5)/8$
& $42.8\pm10.1$ \\

LSTM & 3--7
& $0.560\pm0.011$
& $0.898\pm0.019$
& $0.498\pm0.075$
& $(6.6\pm0.9)/8$
& $52.8\pm5.0$ \\

LSTM & 5--10
& $0.284\pm0.026$
& $0.812\pm0.016$
& $0.234\pm0.039$
& $(6.2\pm0.4)/8$
& $57.4\pm2.4$ \\
\midrule

TCN & 1--3
& $0.480\pm0.022$
& $0.893\pm0.007$
& $0.372\pm0.053$
& $(6.2\pm0.4)/8$
& $50.2\pm0.4$ \\

TCN & 3--7
& $0.515\pm0.023$
& $0.880\pm0.002$
& $0.387\pm0.005$
& $(6.2\pm0.4)/8$
& $52.4\pm1.9$ \\

TCN & 5--10
& $0.310\pm0.035$
& $0.808\pm0.004$
& $0.206\pm0.003$
& $(6.6\pm0.5)/8$
& $59.2\pm4.8$ \\
\bottomrule
\end{tabular}
\end{table*}

The 3--7-day horizon achieved the highest F1-score for all four model families. It also produced the strongest AUROC and PR-AUC for XGBoost and LSTM while providing substantially higher event coverage than the 1--3-day horizon. These results support the 3--7-day interval as the most favorable overall operating point for the present dataset.

The 1--3-day horizon yielded fewer detected events, particularly for logistic regression and XGBoost. Although it provides a warning closer to the event date, the narrower target interval produces fewer positive training samples and appears more difficult to distinguish reliably. TCN was less affected in event coverage, but its F1 and PR-AUC remained below those obtained with the 3--7-day horizon.

The 5--10-day horizon retained relatively high event coverage for the sequence models, but both LSTM and TCN showed substantial reductions in F1 and PR-AUC together with increased alert burden. Broader and more distant warning intervals can therefore preserve event coverage through more diffuse alert streams without providing stronger discrimination.

Overall, the 3--7-day horizon provided the most balanced trade-off among sample-level discrimination, event coverage, and alert burden. This result supports its use as the prespecified reference horizon, while not establishing it as a universally optimal agronomic interval.

\subsection{Sensitivity to the Disease-Free Gap}

The disease-free gap $g$ determines whether a positive annotation is treated as the beginning of a new event or as a continuation of a nearby episode. Changing $g$ therefore changes the event ground truth itself rather than merely modifying a model hyperparameter. The 2023 test year contains nine events for $g=3$, eight events for $g=5$, and only three events for $g=7$. Table~\ref{tab:gap_sensitivity} summarizes the results obtained under the different gap values, which consequently represent related but non-identical prediction tasks.

\begin{table*}[t]
\centering
\caption{Sensitivity to the disease-free event-separation gap on the 2023 test year using the combined E+C representation, $L=30$, and a 3--7-day prediction horizon. Because changing $g$ changes the number and timing of the reference events, detected-event counts are reported relative to the corresponding number of test events. Results for stochastic models are mean $\pm$ standard deviation over five seeds.}
\label{tab:gap_sensitivity}

\small
\setlength{\tabcolsep}{4.2pt}
\renewcommand{\arraystretch}{1.08}

\begin{tabular}{lcccccc}
\toprule
Model &
$g$ &
\makecell{Test\\events} &
F1 &
PR-AUC &
\makecell{Detected\\events} &
Alerts \\
\midrule

LogReg & 3
& 9
& 0.412
& 0.347
& 6/9
& 63 \\

LogReg & 5
& 8
& 0.405
& 0.307
& 6/8
& 46 \\

LogReg & 7
& 3
& 0.188
& 0.097
& 2/3
& 49 \\
\midrule

XGBoost & 3
& 9
& $0.552\pm0.022$
& $0.368\pm0.002$
& $(6.0\pm0.0)/9$
& $40.6\pm0.9$ \\

XGBoost & 5
& 8
& $0.555\pm0.028$
& $0.350\pm0.016$
& $(6.0\pm0.0)/8$
& $41.2\pm1.9$ \\

XGBoost & 7
& 3
& $0.317\pm0.039$
& $0.224\pm0.015$
& $(2.0\pm0.0)/3$
& $36.0\pm5.8$ \\
\midrule

LSTM & 3
& 9
& $0.572\pm0.011$
& $0.487\pm0.066$
& $(7.0\pm0.0)/9$
& $50.0\pm1.6$ \\

LSTM & 5
& 8
& $0.560\pm0.011$
& $0.498\pm0.075$
& $(6.6\pm0.9)/8$
& $52.8\pm5.0$ \\

LSTM & 7
& 3
& $0.262\pm0.100$
& $0.184\pm0.048$
& $(1.8\pm0.4)/3$
& $44.6\pm7.9$ \\
\midrule

TCN & 3
& 9
& $0.550\pm0.015$
& $0.390\pm0.013$
& $(7.2\pm0.4)/9$
& $53.2\pm3.4$ \\

TCN & 5
& 8
& $0.515\pm0.023$
& $0.387\pm0.005$
& $(6.2\pm0.4)/8$
& $52.4\pm1.9$ \\

TCN & 7
& 3
& $0.201\pm0.042$
& $0.125\pm0.002$
& $(1.6\pm0.5)/3$
& $57.2\pm11.6$ \\
\bottomrule
\end{tabular}
\end{table*}

With $g=3$, shorter disease-free interruptions are treated as separate events, increasing the test-event count from eight to nine. LSTM and TCN achieved their highest F1-scores under this definition and detected approximately seven of the nine events. XGBoost performance was similar for $g=3$ and $g=5$, while logistic regression showed only a small change.

The reference gap $g=5$ produced broadly comparable discrimination while retaining eight test events. XGBoost achieved its highest F1 under this setting, while LSTM achieved its highest PR-AUC. With $g=7$, only three test events remained. F1 and PR-AUC declined for all models, and event coverage fell to approximately two events. The small denominator also made the results highly sensitive to the outcome of individual events.

The gap rule is therefore not a minor preprocessing choice. It materially changes the number, timing, and difficulty of the target events. The reference value $g=5$ is retained as a moderate operational episode-separation rule, but it should not be interpreted as an incubation-related or pathogen-specific biological constant.

\subsection{Robustness to Disease-Target Construction}

To assess whether the findings depend materially on the small number of discordant downy- and powdery-mildew annotations, the E+C experiments were repeated using the unified OR target, the consensus AND target, and the separate downy- and powdery-mildew annotations. Logistic regression and XGBoost were used because the objective was to evaluate sensitivity to target construction rather than repeat the complete model-family comparison. Table~\ref{tab:target_robustness} summarizes these experiments, with all other settings held at $L=30$, $g=5$, and a 3--7-day prediction horizon.

\begin{table*}[t]
\centering
\caption{Robustness to alternative disease-target constructions on the 2023 test year using the combined E+C representation, $L=30$, $g=5$, and a 3--7-day prediction horizon. All target constructions contain eight test events. XGBoost results are mean $\pm$ standard deviation over five seeds.}
\label{tab:target_robustness}

\small
\setlength{\tabcolsep}{4pt}
\renewcommand{\arraystretch}{1.08}

\begin{tabular}{llcccc}
\toprule
Target &
Model &
F1 &
AUROC &
PR-AUC &
\makecell{Detected\\events} \\
\midrule

OR & LogReg
& 0.405
& 0.732
& 0.307
& 6/8 \\

OR & XGBoost
& $0.555\pm0.028$
& $0.755\pm0.011$
& $0.350\pm0.016$
& $(6.0\pm0.0)/8$ \\
\midrule

AND & LogReg
& 0.394
& 0.748
& 0.321
& 6/8 \\

AND & XGBoost
& $0.541\pm0.041$
& $0.791\pm0.012$
& $0.465\pm0.025$
& $(5.8\pm0.4)/8$ \\
\midrule

Downy & LogReg
& 0.394
& 0.754
& 0.323
& 6/8 \\

Downy & XGBoost
& $0.573\pm0.038$
& $0.788\pm0.004$
& $0.458\pm0.013$
& $(6.0\pm0.0)/8$ \\
\midrule

Powdery & LogReg
& 0.373
& 0.730
& 0.311
& 6/8 \\

Powdery & XGBoost
& $0.509\pm0.051$
& $0.758\pm0.008$
& $0.353\pm0.019$
& $(5.6\pm0.9)/8$ \\
\bottomrule
\end{tabular}
\end{table*}

The four target constructions produced broadly similar results. Logistic-regression F1 ranged from 0.373 to 0.405, and every target construction resulted in detection of 6 of the 8 test events. XGBoost F1 ranged from $0.509\pm0.051$ for the powdery target to $0.573\pm0.038$ for the downy target, while mean event coverage varied only from $(5.6\pm0.9)/8$ to $(6.0\pm0.0)/8$.

The consensus AND and separate downy targets produced higher XGBoost AUROC and PR-AUC than the unified OR target. These differences should not be interpreted as evidence that one pathogen is intrinsically more predictable. The source annotations differ on very few training days and are identical throughout the validation and test years. The performance differences therefore arise primarily from minor changes in the training labels rather than from distinct pathogen-specific test events.

The principal conclusions are consequently not an artifact of the logical OR operation. At the same time, the near-equivalence of the four constructions confirms that the dataset contains insufficient independent pathogen-specific variation to support reliable comparison of downy- and powdery-mildew warning mechanisms.

\subsection{Interpretation of Seasonal and Environmental Information}

Global SHAP analysis was used to examine how the representative seed-0 XGBoost model distributed its attribution across the combined E+C inputs. In cumulative terms, humidity-related variables accounted for approximately 34.3\% of total absolute attribution, temperature-related variables for 28.7\%, calendar variables for 21.1\%, and rainfall-related variables for 16.0\%. The missing-day indicator received negligible attribution.

These cumulative percentages are affected by the different numbers of feature--lag columns in each group. After normalization by group size, the attribution indices were 1.37 for calendar, 1.11 for humidity, 1.07 for temperature, and 0.69 for rainfall, where 1 denotes the average attribution per feature--lag column. Thus, although humidity received the largest cumulative attribution, calendar variables received the strongest average attribution after accounting for the number of predictors in each group.

Temporal aggregation assigned 69.3\% of total absolute attribution to observations 14--29 days before the prediction date. The remaining attribution was distributed across lags 7--13 days (14.3\%), 0--2 days (9.3\%), and 3--6 days (7.1\%). After accounting for interval width, the corresponding attribution indices were 0.93, 0.53, 0.61, and 1.30 for the 0--2-, 3--6-, 7--13-, and 14--29-day intervals, respectively. The oldest interval therefore retained above-average attribution even after controlling for its greater number of lag positions. However, this pattern may reflect longer-range seasonal context, redundancy among lagged descriptors, or recurring management-related timing rather than a biological disease precursor.

The SHAP findings should be considered together with the input-representation experiments. Although the combined model used meteorological variables, environmental-only models were substantially weaker than calendar-only models, and adding environmental variables did not produce consistent improvements across model families. Consequently, the analysis does not support attributing model performance to a single environmental precursor. It instead indicates that the fitted predictions use a mixture of seasonal and meteorological information.

\subsection{Operational Interpretation and Limitations}

The experiments do not support the conclusion that sequence models consistently identify environmental disease precursors more effectively than simpler alternatives. Calendar-only models were highly competitive, environmental-only models were substantially weaker, and the contribution of environmental variables beyond calendar timing varied across model families. Model rankings also changed with the input representation, historical window, prediction horizon, and event-separation gap. The principal findings therefore concern target formulation, seasonal confounding, and operational evaluation rather than the superiority of a particular architecture.

The comparison with conventional daily-status classification further shows that event-onset prediction should not be presented as uniformly numerically superior. Daily-status logistic regression detected one more event than its event-onset counterpart under the common warning protocol, but generated more alerts. Event-onset XGBoost detected one more event than daily-status XGBoost, but also generated more alerts. The two formulations therefore produce different trade-offs. The methodological advantage of the event-onset formulation is that its target directly represents the intended question of whether a new episode will begin within a specified warning horizon.

The seasonal baseline and input-representation experiments demonstrate that treatment-derived targets require explicit calendar controls. Monthly climatology alone detected 6 of 8 events, while the strongest calendar-only models detected 7 of 8 and frequently matched or exceeded the corresponding combined configurations. SHAP analysis showed that the representative combined XGBoost model used both calendar and meteorological variables, but attribution magnitude does not establish independent predictive contribution or biological causality. These findings do not prove that treatment decisions followed fixed calendar schedules, but they show that seasonal timing and unrecorded management-related factors may contribute substantially to the annotations.

The event-level results must be interpreted cautiously. The 2023 test year contains only eight reference events, so the outcome of one event changes event recall by 0.125. For illustration, recall values of $7/8$ and $6/8$ have approximate 95\% Wilson confidence intervals of $[0.529,0.978]$ and $[0.409,0.929]$, respectively, which overlap substantially. Random-seed repetition quantifies optimization variation but does not increase the number of independent test events. Differences of one event therefore do not establish statistical superiority.

Changing the historical window or prediction horizon alters the number and composition of positive samples, while changing the disease-free gap also changes the event ground truth. The reference configuration should consequently be viewed as an operational formulation that performed reasonably in this case study rather than as a universally optimal biological specification.

The downy- and powdery-mildew annotations originate from fungicide-treatment records, show 98.9\% daily agreement, and are identical in the validation and test years. The resulting any-mildew target is therefore an operational management-risk proxy rather than a pathogen-specific observation of infection onset or disease severity. The target-robustness experiments show that the principal findings are not driven by the logical OR operation, but the near-equivalence of the alternative targets prevents meaningful pathogen-specific interpretation.

Finally, the study is based on one vineyard, four annual cycles, and a single future test season. The chronological split provides a realistic assessment of temporal transfer within this site, but it does not establish generalization across vineyards, cultivars, climates, sensing systems, or management practices. Broader validation using additional sites, more independent seasons, and pathologically confirmed pathogen-specific observations is required before the framework can be interpreted as a biological disease-onset warning system.

Overall, forecasting treatment-associated vineyard risk annotations is shaped by target construction, recurring seasonal structure, model choice, and operational evaluation criteria. Event-oriented assessment remains valuable because it makes onset coverage, warning timing, and alert burden explicit. At the same time, management-derived proxy targets require seasonal baselines, careful causal interpretation, and restrained claims regarding biological disease mechanisms.

\section{Conclusion and Future Work}

This study investigated an event-based formulation for short-horizon warning of treatment-associated vineyard mildew-risk episodes from agro-meteorological time series. Rather than classifying a persistent daily status, the primary task was defined as forecasting whether a new gap-separated risk event would begin within the following 3--7 days. The framework combined causal preprocessing, historical environmental windows, cyclic calendar encoding, several established model families, and an event-oriented evaluation protocol covering event coverage, warning lead time, alert burden, and strict false-alert episodes.

The results show that the principal contribution of the study lies in target formulation and evaluation rather than in the superiority of a particular predictive architecture. Calendar-only models were highly competitive and frequently matched or exceeded models using environmental variables, whereas environmental-only representations were substantially weaker. A monthly climatological baseline detected 6 of the 8 test events, while the strongest calendar-only models detected 7 of 8. Adding agro-meteorological variables to calendar information produced inconsistent changes across model families. These findings indicate that the treatment-derived target contains substantial recurring seasonal structure and that strong warning performance cannot be attributed exclusively to short-term environmental conditions.

The comparison with conventional daily-status classification further clarified the role of the event-onset formulation. The two tasks produced different trade-offs rather than uniform superiority of either formulation. Daily-status logistic regression detected one more event than its event-onset counterpart under the common warning protocol, but generated more alerts. Event-onset XGBoost detected one more event than daily-status XGBoost, while also generating more alert days. The value of the event formulation therefore lies not in guaranteed numerical improvement, but in aligning the supervised target directly with the operational objective of warning before a new episode and in making event coverage, warning timing, and alert burden explicit.

The sensitivity analyses showed that the conclusions also depend on the temporal definition of the task. The 3--7-day horizon provided the most balanced overall performance across the evaluated models, but no historical window length was uniformly optimal. The disease-free gap had an even stronger influence because changing it altered the number and timing of the reference events. These parameters should therefore be treated as operational target-construction choices rather than fixed biological constants.

The SHAP analysis supports a similarly cautious reading. The representative E+C XGBoost model assigned attribution to both calendar and meteorological variables, with substantial attribution also assigned to older historical lags. However, SHAP describes conditional model usage rather than independent predictive contribution or biological causality. Considered together with the strong calendar-only performance and inconsistent gains from adding environmental variables, the results do not support the identification of a single biological disease precursor.

Several limitations remain. The study is based on a single vineyard, four annual cycles, and only eight reference events in the final test year. Repetition across random seeds quantifies optimization variability but does not overcome the uncertainty associated with the small number of independent events. Moreover, the labels originate from fungicide-treatment records rather than direct pathological confirmation, and the downy- and powdery-mildew annotations contain very limited pathogen-specific variation. The findings should therefore be interpreted as forecasting an operational any-mildew management-risk proxy rather than directly predicting biological infection onset, symptom appearance, or disease severity.

Future work should evaluate the formulation across additional vineyards, cultivars, climates, and seasons using pathogen-specific and pathologically confirmed observations. Richer annotation protocols should distinguish treatment timing, visible symptoms, infection indicators, phenological stage, expert assessments, and operational constraints, allowing environmental risk and management behavior to be analysed separately. Multimodal extensions combining agro-meteorological measurements with imaging, leaf-wetness sensing, phenological information, or mechanistic infection models may provide more biologically grounded warning signals. Prospective evaluation should also examine probability calibration, cost-sensitive thresholds, uncertainty-aware warning policies, and real-time performance on edge and IoT platforms deployed in vineyard environments.

Overall, this study demonstrates that event-oriented evaluation provides a more transparent account of short-horizon warning behavior than point-wise daily classification alone. It also shows that management-derived agricultural targets may contain substantial seasonal and operational structure. Explicit seasonal baselines, careful target construction, and restrained biological interpretation are therefore necessary to ensure that reported model performance reflects genuine anticipatory value rather than recurring calendar patterns or properties of the annotation process.

\section*{Funding}
The authors received no financial support for the research, authorship, and/or publication of this article.

\bibliography{sn-bibliography}

\end{document}